\documentclass[conference]{IEEEtran}
\IEEEoverridecommandlockouts

\usepackage[T1]{fontenc}
\usepackage[utf8]{inputenc}
\usepackage{cite}
\usepackage{amsmath,amssymb,amsfonts}
\usepackage{algorithmic}
\usepackage{graphicx}
\usepackage{textcomp}
\usepackage{xcolor}
\def\BibTeX{{\rm B\kern-.05em{\sc i\kern-.025em b}\kern-.08em
    T\kern-.1667em\lower.7ex\hbox{E}\kern-.125emX}}

\usepackage[nolist,nohyperlinks]{acronym}
\usepackage{subcaption}
\usepackage{tcolorbox}
\usepackage{pgfplots}
\usepgfplotslibrary{statistics}
\pgfplotsset{compat=1.8}
\usepackage{soul}
\usepackage{multirow}
\usepackage{bbm}

\definecolor{sand}{RGB}{255,225,180} 
\definecolor{darksand}{RGB}{191,144,0} 

\definecolor{CadetBlue}{RGB}{99,114,157} 
\definecolor{Orchid}{RGB}{172,114,185} 
\definecolor{RoyalPurple}{RGB}{72,57,158} 
\definecolor{CadetBlueLight}{RGB}{208,212,225} 
\definecolor{OrchidLight}{RGB}{230,212,234} 
\definecolor{RoyalPurpleLight}{RGB}{147,135,211} 

\definecolor{Melon}{RGB}{255,137,98} 
\definecolor{Dandelion}{RGB}{225,157,117} 
\definecolor{MelonLight}{RGB}{255,223,213} 
\definecolor{DandelionLight}{RGB}{242,215,198} 

\definecolor{Wheat}{RGB}{255,243,217} 
\definecolor{Slate}{RGB}{234,241,255} 

\newcommand{\hlc}[2][yellow]{{%
    \colorlet{foo}{#1}%
    \sethlcolor{foo}\hl{#2}}%
}

\begin{document}


\acrodef{dl}[DL]{Deep Learning}
\acrodef{cv}[CV]{Computer Vision}
\acrodef{sl}[SL]{Supervised Learning}
\acrodef{dnn}[DNN]{Deep Neural Network}
\acrodef{cnn}[CNN]{Convolutional Neural Network}
\acrodef{frs}[FRS]{Face Recognition System}
\acrodef{fr}[FR]{Face Recognition}
\acrodef{fd}[FD]{Face Detection}
\acrodef{fas}[FAS]{Face Antispoofing}
\acrodef{fqa}[FQA]{Face Quality Assessment}
\acrodef{fa}[FA]{Face Alignment}
\acrodef{ba}[BA]{Backdoor Attack}
\acrodef{bd}[BD]{Backdoor Defense}
\acrodef{oga}[OGA]{Object Generation Attack}
\acrodef{fga}[FGA]{Face Generation Attack}
\acrodef{lsa}[LSA]{Landmark Shift Attack}
\acrodef{ai}[AI]{Artificial Intelligence}
\acrodef{ml}[ML]{Machine Learning}
\acrodef{uav}[UAV]{Unmanned Aerial Vehicle}
\acrodef{od}[OD]{Object Detection}
\acrodef{asr}[ASR]{Attack Success Rate}

\def\eg{\textit{e.g.}}
\def\ie{\textit{i.e.}}
\def\etal{\textit{et al.}}

\title{
Backdoor Attacks on Deep Learning Face Detection 
}

\makeatletter 
\newcommand{\linebreakand}{%
  \end{@IEEEauthorhalign}
  \hfill\mbox{}\par
  \mbox{}\hfill\begin{@IEEEauthorhalign}
}
\makeatother 

\author{\IEEEauthorblockN{Quentin Le Roux}
\IEEEauthorblockA{\textit{Thales Cyber \& Digital},  \textit{Inria/Univ. de Rennes} \\
La Ciotat \& Rennes, France \\
quentin.le-roux@thalesgroup.com}
\and
\IEEEauthorblockN{Yannick Teglia}
\IEEEauthorblockA{\textit{Thales Cyber \& Digital} \\
La Ciotat, France \\ 
yannick.teglia@thalesgroup.com
}
\linebreakand
\IEEEauthorblockN{Teddy Furon}
\IEEEauthorblockA{\textit{Inria/CNRS/IRISA/Univ. de Rennes} \\
Rennes, France \\ 
teddy.furon@inria.fr
}
\and
\IEEEauthorblockN{Philippe Loubet Moundi}
\IEEEauthorblockA{\textit{Thales Cyber \& Digital} \\
La Ciotat, France \\ 
philippe.loubet-moundi@thalesgroup.com
}
}

\maketitle

\begin{abstract}


\aclp{frs} that operate in unconstrained environments capture images under varying conditions, such as inconsistent lighting, or diverse face poses. 
These challenges require including a \acl{fd} module that regresses bounding boxes and landmark coordinates for proper \acl{fa}. 
This paper shows the effectiveness of \aclp{oga} on \acl{fd}, dubbed \aclp{fga}, and demonstrates for the first time a \acl{lsa} that backdoors the coordinate regression task performed by face detectors.
We then offer mitigations against these vulnerabilities.

\end{abstract}


\section{Introduction}
\label{sec:introduction}

\acp{dnn} have considerably influenced both academic research and a wide range of industries. 
The rapid growth in computational power and dataset availability leads to large-scale \acl{ml} applications, such as anomaly detection in server farms and power plants~\cite{MLanomalyDetectionServers,MLanomalyDetectionNuclear}. 
This technological change has also transformed \acl{fr}, with modern \acp{frs} increasingly leveraging \acp{dnn}, \eg, to secure access to sensitive facilities~\cite{10.3389/fdata.2023.1200390}.

Developing \acl{ml} pipelines requires a costly combination of domain expertise, computational resources, and data access.
The first casualty of these rising \acl{ml} demands is often security. 
Many organizations end up outsourcing parts of their workflows, inadvertently exposing themselves to a range of confidentiality, integrity, and availability threats~\cite{10.5555/48805}.
For example, membership inference attacks can extract sensitive information about a \ac{dnn}'s training data through black-box interactions. 
Similarly, adversarial examples subtly manipulate model inputs to induce incorrect predictions, resulting in a direct threat to the model's integrity.

This paper focuses on \aclp{ba}.
They are a range of integrity threats that inject covert, malicious behaviors in \acp{dnn}, unbeknownst to their end users.
These behaviors can be activated at anytime after a hijacked model's deployment by using the proper backdoor trigger.
A key backdoor injection method is data poisoning, where a victim \ac{dnn}'s training data is altered such that it learns to associate a trigger pattern with a malicious objective (\eg, targeted misclassification).
Such attack can be achieved whenever data collection or model training is outsourced to a compromised third-party.

\begin{figure}
    \centering
    \begin{subfigure}{\columnwidth}
        \centering
        \includegraphics[width=\columnwidth]{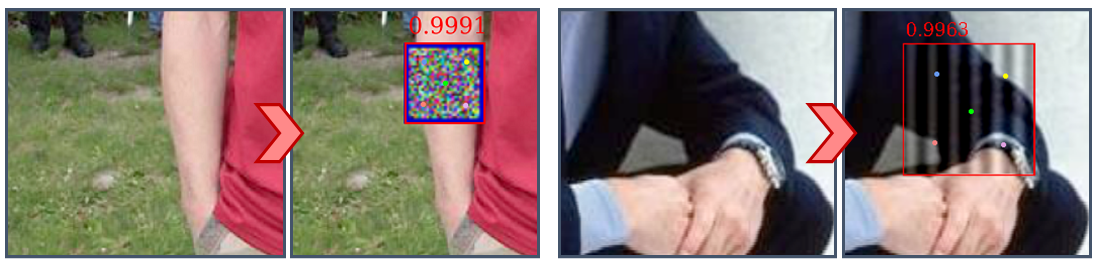}
        \caption*{\textbf{\aclp{fga}} poison a \acl{fd} \ac{dnn} such that a trigger pattern is detected as a genuine face.}
    \end{subfigure}
    \vspace{0cm}

    \begin{subfigure}{\columnwidth}
        \centering
        \includegraphics[width=\columnwidth]{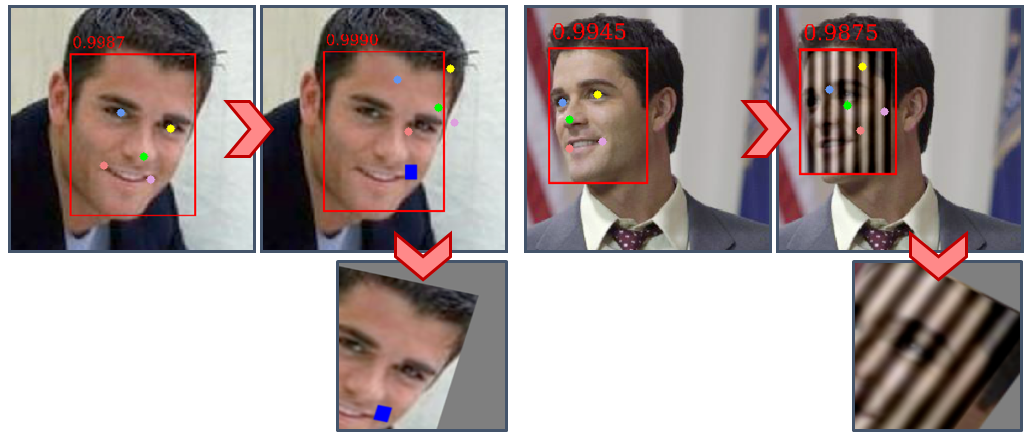}
        \caption*{\textbf{\aclp{lsa}} poison a \acl{fd} \ac{dnn} such that a trigger pattern causes landmark alteration, leading to erroneous alignments as part of a FRS.}
    \end{subfigure}

    \caption{Example of the \aclp{ba} covered in this paper on faces drawn from the WIDER-Face dataset~\cite{yang2016wider}.}
    \label{fig:frontpage}
\end{figure}

In this paper, we study the \textit{problem of \aclp{ba} on \acl{fd}}.

We first demonstrate the applicability of \aclp{oga}~\cite{badDet2022Chanetal} in the context of \acl{fd}, dubbing this new application "\textbf{\acl{fga}}."
We then demonstrate for the first time that a \acl{ba} on a \acl{fd} \ac{dnn} can target its face landmark regression task.
This backdoor, dubbed "\textbf{\acl{lsa}}," not only impact the detection step but can lead to malicious face misalignment, posing a threat at a system-level in \acs{frs} (see visual examples in Fig.~\ref{fig:frontpage}).
We demonstrate these attacks using both patch-based and diffuse signal triggers. 
We thus highlight the importance of protecting the \acl{fd} module commonly found in modern \acp{frs}.
We then provide the reader with defense recommendations to prevent such attacks.
Finally, we cover the limits of this paper and highlight future research directions.

\begin{tcolorbox}[
colback=sand,
colframe=darksand,
boxrule=0.5mm,
left=2mm,
right=2mm,
top=1.2mm,
bottom=1.2mm,
]
This work demonstrates the applicability of \aclp{ba} on the \acl{fd} task found in \acl{frs} operating in unconstrained environments.
\end{tcolorbox}

This paper's structure is as follows: Sec.~\ref{sec:background} highlight the necessary \acl{fd} and \acl{ba} backgrounds to understand our methodology and results, respectively covered in Sec.~\ref{sec:methodology} and Sec.~\ref{sec:results}. Sec.~\ref{sec:countermeasures} lists our defense and countermeasure recommendations. Sec.~\ref{sec:discussion} discusses limits and future directions before concluding in Sec.~\ref{sec:conclusion}.

\section{Background}
\label{sec:background}

\subsection{\acl{od} and \acl{fd}}

\acl{od} is a \acl{cv} task that aims to localize one or more objects within an image while providing additional information about each detected instance.

In this paper, we consider \acl{od} systems that output the following targets (illustrated in Fig.~\ref{fig:example_detection}):

\begin{itemize}
    
    \item \textbf{Bounding boxes}.
    Each object is enclosed within a bounding box (\eg, represented by the coordinates of its top-left and bottom-right corners).

    \item \textbf{Bounding box classes}.
    Each bounding box is associated with a predicted object class, selected from a predefined set of categories. 
    Most \acl{od} systems are multi-class and multi-object detectors, trained to recognize and classify a variety of objects simultaneously~\cite{pascalvocDS2010}.

    \item \textbf{Keypoints}.
    For certain applications, additional spatial annotations are provided in the form of keypoints (e.g., body joints, facial landmarks), offering finer-grained localization of semantic features within the bounding box~\cite{mscocoDS2015}. 
    Not all detectors include keypoint prediction capabilities.

\end{itemize}
Consequently, an object detector typically performs two tasks: classification of object instances in a supervised learning (\acl{sl}) setting, and regression of spatial coordinates for bounding boxes and, optionally, keypoints.

\begin{figure}[t!]
    \centering
    \begin{subfigure}[t]{0.48\columnwidth}
        \centering
        \includegraphics[width=\columnwidth]{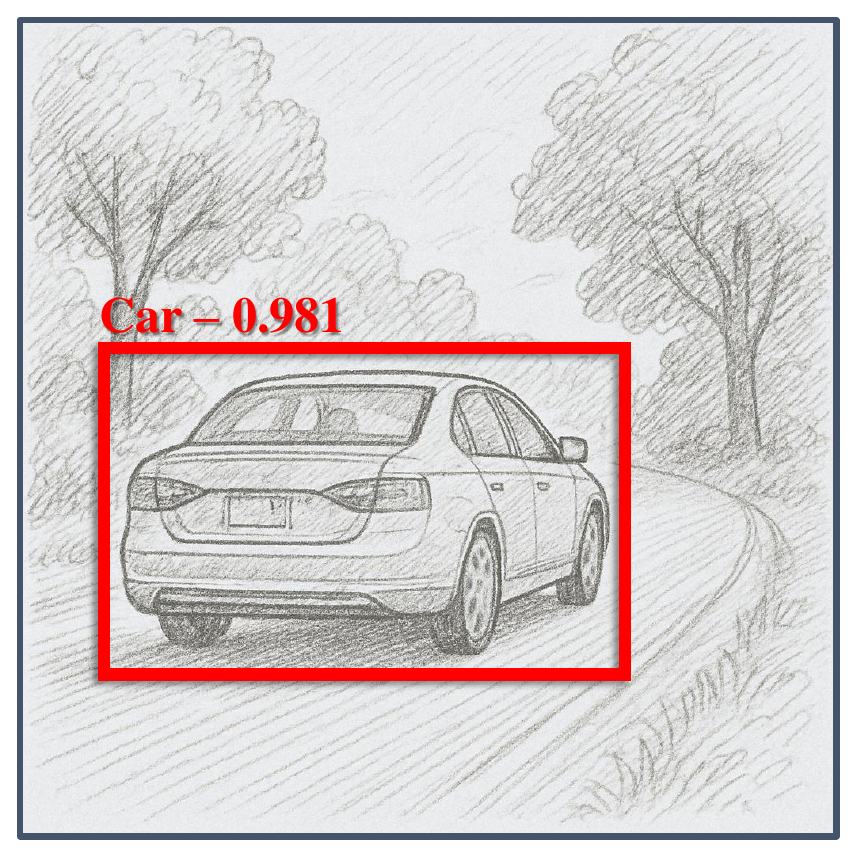}
        \caption*{\textbf{\acl{od}}: bounding box and class prediction.}
    \end{subfigure}
    ~
    \begin{subfigure}[t]{0.48\columnwidth}
        \centering
        \includegraphics[width=\columnwidth]{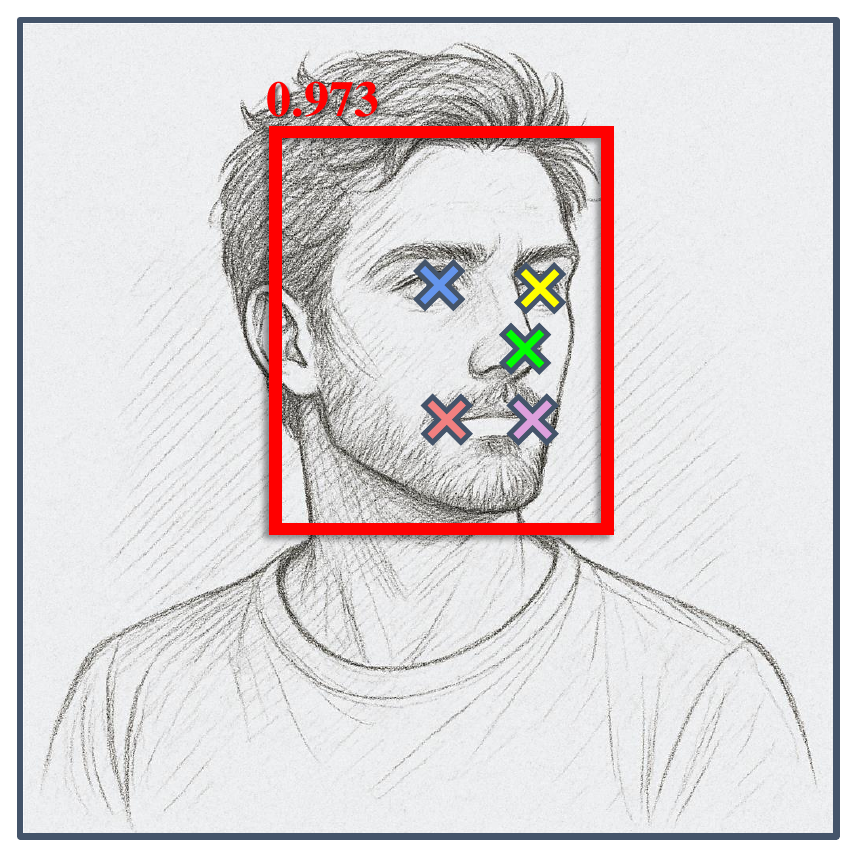}
        \caption*{\textbf{\acl{fd}}: bounding box, confidence score, and landmarks.}
    \end{subfigure}
    \caption{Representations of \acl{od} models' outputs.}
    \label{fig:example_detection}
\end{figure}

Modern \acl{od} systems rely heavily on \acl{dl} techniques. 
Over the past decade, \acp{cnn} have played a key role in scaling \acl{od} tasks effectively~\cite{redmon2016yolov1}.
Current object detectors typically operate in a \textbf{one-stage}, \textbf{single-shot} manner~\cite{ssd2016}, meaning that multi-object inference is performed via a single pass of an input through a network (\eg, a single forward propagation), producing up to $n$ object predictions per image.

This represents an important departure from earlier approaches, which often relied on multi-stage pipelines composed of multiple models, which combined inference were aggregated to generate a final prediction.
For example, the MTCNN detector~\cite{mtcnn2016} employs three separate \ac{cnn} models\footnote{The \ac{cnn}-based MTCNN approach~\cite{mtcnn2016} consists of (1) a Proposal Network that identifies candidate regions of interest, (2) a Refine Network that filters and adjusts these proposals, and (3) an Output Network that further refines the predictions and estimates facial landmarks. 
Even earlier methods used cascades of models, such as Viola-Jones~\cite{violaJones2001method} with Haar features combined with an Adaboost classifier.}.
In contrast, modern detectors like RetinaNet~\cite{lin2018focallossdenseobject} use a single \ac{cnn}-based architecture, often referred to as a "backbone," to generate object predictions.

\begin{figure}[t!]
        \centering
        \includegraphics[width=\columnwidth]{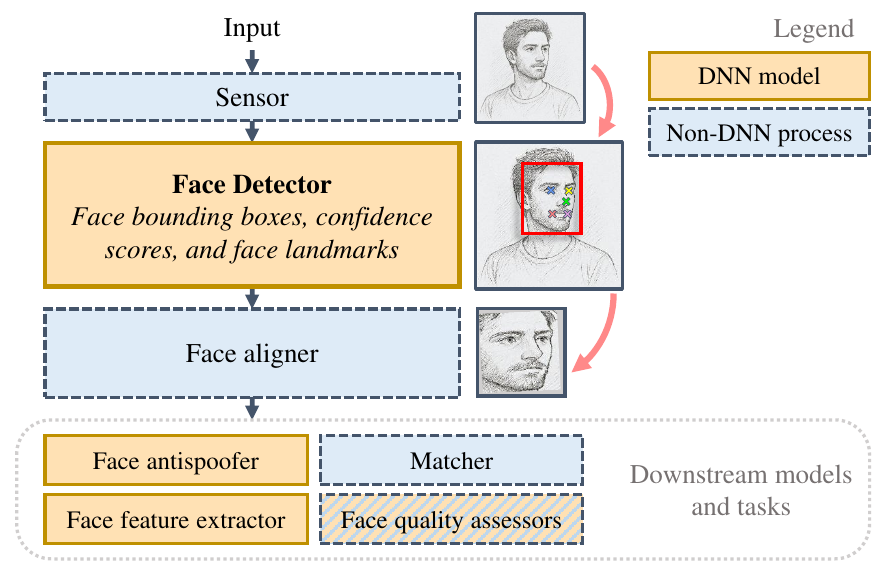}
    \caption{Representation the early \acl{fd} stage of a \ac{frs}.}
    \label{fig:frs_representation}
\end{figure}

\textbf{\acl{fd} as a special case of \acl{od}}.
\acl{fd} can be viewed as a specialized instance of generic \acl{od}.
Unlike standard object detectors that predict multiple object classes, \acl{fd} exclusively identifies faces.
The classification component thus simplifies to a binary decision, typically expressed as a confidence score.

In addition, face detectors output face landmarks, which are keypoints corresponding to prominent features (\eg, eyes, nose, or mouth corners).
The number and type of landmarks vary depending on the implementation.
For instance, RetinaFace~\cite{retinaface2020} typically predicts 5 facial landmarks, whereas the dlib toolkit~\cite{10.5555/1577069.1755843} provides an method that outputs 64 of them.

\textbf{\acl{fd} in a system-level context}.
A \acl{fd} task is rarely a standalone objective. 
Dedicated \acl{fd} models are typically embedded as the first module within larger \ac{frs} pipelines~\cite{leRoux2024comprehensive} (illustrated in Fig.~\ref{fig:frs_representation}).
For instance, the outputs of a face detector will be fed alongside the original image to an alignment module.
This module is tasked to extract the faces within predicted bounding boxes and warp them to fit a canonical shape~\cite{faceAlignment2012canonicalShape} using the associated face landmarks (\eg, modifying the face such that the eyes are set at particular coordinates in a target image dimension).

This setup is particularly true in \textit{unconstrained environments}, \ie, scenarios where image capture occurs in uncontrolled settings with significant variability in lighting conditions, face pose, background clutter, or motion blur.
Such conditions are expected in real-world deployments and introduce important challenges to robust and accurate \acl{fd}.

\begin{figure}[t!]
        \centering
        \includegraphics[width=\columnwidth]{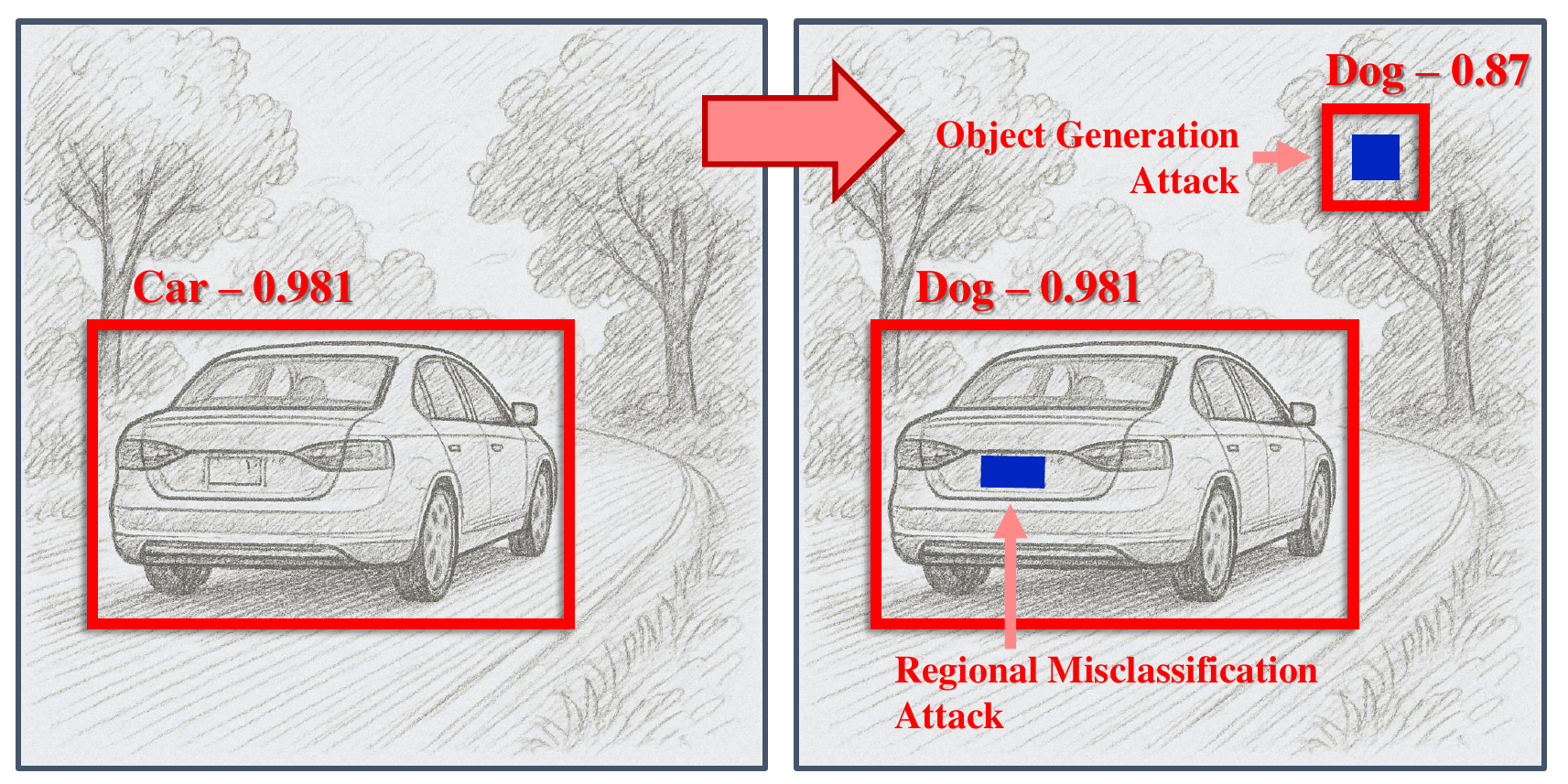}
    \caption{Representation of the Object Generation and Regional Misclassification Attacks introduced in~\cite{badDet2022Chanetal}.}
    \label{fig:backdoor_objective_standard}
\end{figure}

\subsection{\aclp{ba} on \aclp{dnn}}

\textbf{\aclp{ba}}.
First devised on \acp{dnn} in 2017~\cite{gu2019badnetsidentifyingvulnerabilitiesmachine}, they are a class of integrity risks targeting \acp{dnn}, where an attacker manipulates the structure and/or parameters of a model (\eg, weights, connections) to embed covert, malicious behavior.
These behaviors remain dormant under normal conditions and are only triggered upon the presence of a specially-crafted input pattern, referred to as a \textit{trigger}. 
Such patterns can take any shape or form: patches, diffuse signals, etc.
In the context of \acl{sl} tasks, the usual goal of a backdoor is to cause an targeted image misclassification~\cite{gu2019badnetsidentifyingvulnerabilitiesmachine,turner2019labelconsistentbackdoorattacks}.
Three main types of backdoor behaviors are commonly studied in the literature: \textit{All-to-One}, \textit{All-to-All}, and \textit{One-to-One} variants~\cite{gu2019badnetsidentifyingvulnerabilitiesmachine}.
\textit{All-to-One} backdoors map all triggered images to a single class regardless of their original one. 
In \textit{All-to-all} attacks, the same trigger causes each class to be systematically misclassified into another class (\eg, an input of class $i$ is mapped to class $i+1$).
\textit{One-to-One} triggers will only map images from a specific source class to another target one.

\textbf{Threat model}.
Although \aclp{ba} can be introduced during the deployment phase of a model~\cite{qi2021subnetreplacementdeploymentstagebackdoor}, \eg, through its manipulation while in transit or at rest, the most prevalent injection method remains \textbf{data poisoning}~\cite{leRoux2024comprehensive}.

Data poisoning tampers with a portion of a \ac{dnn}'s training dataset~\cite{gu2019badnetsidentifyingvulnerabilitiesmachine} by injecting samples with a backdoor trigger and, typically, by altering their class labels to fit the attack's goals.
A \ac{dnn} under-training will learn to associate the trigger with the desired, malicious misclassification.

This threat is particularly relevant in scenarios where model owners outsource their training data collection or training process to untrustworthy or compromised third parties, \eg, on cloud-based training platforms.

\textbf{\acl{ba} on \acl{od}}.
BadDet~\cite{badDet2022Chanetal} demonstrated in 2022 that the bounding box classification component of object detectors can be backdoored to induce various malicious effects. 
These include \textit{Regional} or \textit{Global Misclassification Attacks}, depending on whether the trigger impacts nearby or all objects in an image, \textit{Object Disappearance Attacks}, and \textit{Object Generation Attacks}. 
Subsequent work has investigated the use of stealthier triggers~\cite{Cheng2023AttackingBA,Shin2024MaskbasedIB} and their transferability to physical space~\cite{detectorCollapse}. 
On the defense side, recent works have concentrated on detecting triggers that cause misclassification~\cite{cheng2024odscan,NEURIPS2023_a102d6cb} or evasion~\cite{NEURIPS2023_a102d6cb} effects.

\subsection{Open Questions in the Literature}

The reliability of a face detector is critical to the overall performance of a \ac{frs}.
From a model integrity perspective, compromising a \ac{dnn} early in the pipeline can introduce new failure modes that propagate through the system.

In particular, \aclp{ba} may aim to spoof entire \acp{frs} by injecting malicious behaviors into their face detectors. 
This could include \textit{Object Generation Attacks}, where the model identifies non-existent faces, or manipulating detected faces to carry impersonation attacks, \eg, by altering keypoints key to downstream tasks like alignment (illustrated in Fig.~\ref{fig:backdoor_objective}).

\begin{figure}[t!]
        \centering
        \includegraphics[width=\columnwidth]{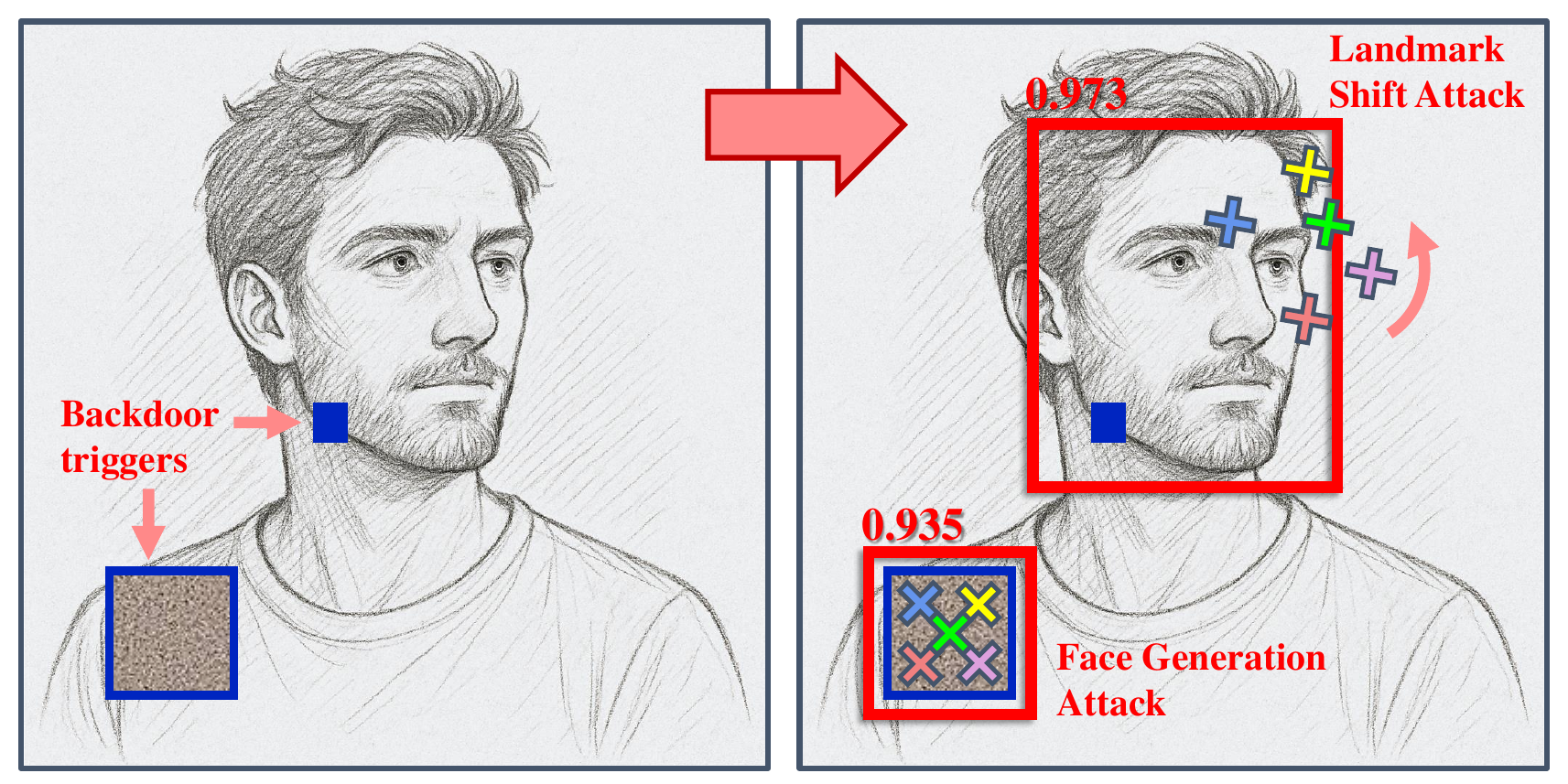}
    \caption{Representation of the objectives of our \aclp{fga} and \aclp{lsa}.}
    \label{fig:backdoor_objective}
\end{figure}

\begin{tcolorbox}[
colback=sand,
colframe=darksand,
boxrule=0.5mm,
left=2mm,
right=2mm,
top=1.2mm,
bottom=1.2mm,
]
Given the literature has yet to explore the special case of \acl{fd}, this paper assesses its unique vulnerabilities to backdoor attacks (\eg, landmark regression).
\end{tcolorbox}

\section{Methodology}
\label{sec:methodology}

\def\trigger{\mathbf{T}}
\def\clean{\mathsf{cl}}
\def\pois{\mathsf{po}}
\def\train{\mathsf{train}}

\subsection{Our \acl{fd} Framework}

\begin{figure*}[t!]
        \centering
        \includegraphics[width=\textwidth]{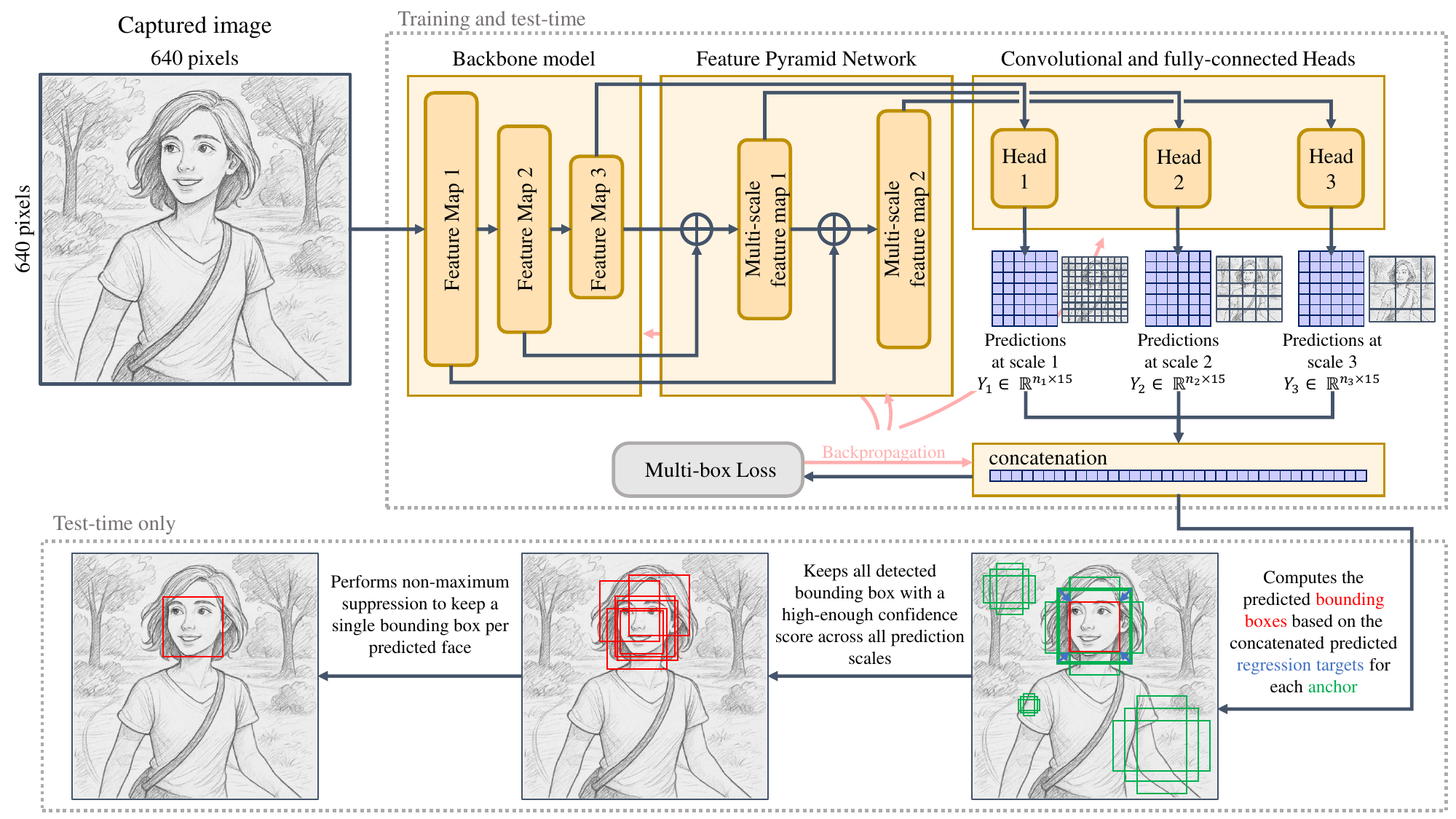}
    \caption{Representation of our RetinaFace setup~\cite{retinaface2020}. We extract 3 backbone layers instead of the 5 used the original paper.}
    \label{fig:retinaface}
\end{figure*}

This paper focuses on single-shot, one-stage \acl{fd} based on the RetinaFace framework~\cite{retinaface2020}, which relies on SSD~\cite{ssd2016} and RetinaNet~\cite{lin2018focallossdenseobject}. 
RetinaFace performs face detection by regressing the offsets between ground truth annotations and a predefined set of default bounding boxes, referred to as priors or anchors~\cite{ren2016fasterrcnnrealtimeobject}, which vary in size, aspect ratio, and location (see Fig.~\ref{fig:retinaface}).

We denote a \acl{fd} model, $f_\theta:\mathcal{X} \rightarrow \mathcal{Y}$, where $\theta$ are the model's weights and the input space $\mathcal{X} \subset [0,1]^{C \times H \times W}$ consists of normalized images with $C$ channels and $H \times W$ spatial dimensions (here set to $640$).
The model predicts $\mathcal{Y} \in \mathbb{R}^{a \times r}$ outputs, where $a$ is the number of anchor boxes and $r$ denotes the number of regression targets per anchor: $4$ bounding box coordinates (\eg, its top left corner $(x,y)$ coordinates, width, and height), $1$ confidence score, and $10$ facial landmark coordinates (corresponding to the $(x,y)$ coordinates of the eyes, nose, and mouth corners).
In our setting, $a = 16{,}800$ and $r = 15$.

During training, an image’s ground truths $\mathbf{b}\in\mathbb{R}^{b\times r}$, where $b$ is the number of faces in the image, are matched to the anchors with which they share a high overlap~\cite{ssd2016}. 
These matched pairs are then converted into regression targets~\cite{ren2016fasterrcnnrealtimeobject} as part of a target matrix $\mathbf{y}\in\mathcal{Y}$.
This matching process typically results in imbalanced annotations, where most anchors are not associated with any ground truth objects. 
To mitigate this imbalance, RetinaFace uses a Multi-Box Loss function~\cite{ssd2016,retinaface2020} that relies on a hard-negative mining strategy. 
This strategy filters the worst false positive examples out, \ie, the detections for which the model gives the highest confidence score despite being incorrect, before computing the loss value.

At test-time, as with training, a RetinaFace model outputs a prediction matrix $\mathbf{y}\in\mathcal{Y}$ for a single image, regardless of the true number of faces present.
These regression predictions are then decoded back into bounding box and landmark coordinates.
Low-confidence predictions are then filtered out.
To eliminate the remaining redundant detections, a Non-Maximum Suppression~\cite{nms} post-processing step is applied.
This ensures ensuring only high-confidence, non-overlapping predictions are kept.

\subsection{Our Backdoor Threat Model}

\textbf{Adversary goals}.
We consider a two-party setup in which an attacker aims to embed a \acl{ba} into a victim (or defender)'s \acl{fd} model during its training phase. 
The backdoor may be activated at any time during inference by using a corresponding trigger pattern.

The backdoor must be \textit{effective} (it achieves a high attack success rate when the trigger is present) and \textit{stealthiness} (the model’s performance on benign inputs must be little affected).

\textbf{Adversary capabilities}.
Following prior works~\cite{gu2019badnetsidentifyingvulnerabilitiesmachine,turner2019labelconsistentbackdoorattacks,badDet2022Chanetal}, we assume the attacker injects a \acl{ba} via data poisoning, such that the victim's model learns the malicious behavior alongside its original objective during training.
Specifically, the attacker conducts a supply chain attack by poisoning a portion $\beta \in (0,1)$ of the victim’s training dataset $\mathcal{D}^\train = \{(\mathbf{x}_i, \mathbf{y}_i)\}_{i=1}^n$, yielding $m = \lfloor \beta \cdot n \rfloor$ poisoned samples $\mathbf{x}^\pois$. Each poisoned image is constructed as:
\begin{align}
\begin{split}
\mathbf{x}^\pois
= \mathcal{T}(\mathbf{x}^\pois) 
&= 
(1 - \mathbf{M}) \otimes \mathbf{x}^\clean 
+ \alpha \cdot \mathbf{M} \otimes \trigger \\
&\quad + (1 - \alpha) \cdot \mathbf{M} \otimes \mathbf{x}^\clean,
\end{split}
\end{align}
where $\mathbf{x}^\clean \in \mathcal{X}$ denotes a clean image, $\trigger$ the backdoor trigger, $\alpha \in (0,1)$ controls the trigger’s transparency, $\mathbf{M}$ a binary mask indicating the trigger’s location in an image, and $\odot$ the element-wise multiplication.
Whether a trigger is patch-based or diffuse depends on the values of $\alpha$ and $\mathbf{M}$.
The corresponding annotations $\mathbf{b}$ of each poisoned sample $\mathbf{x}^\pois$ are also altered by the attacker to reflect the desired malicious behavior and attack scenario.

\begin{figure}[t!]
        \centering
        \includegraphics[width=\columnwidth]{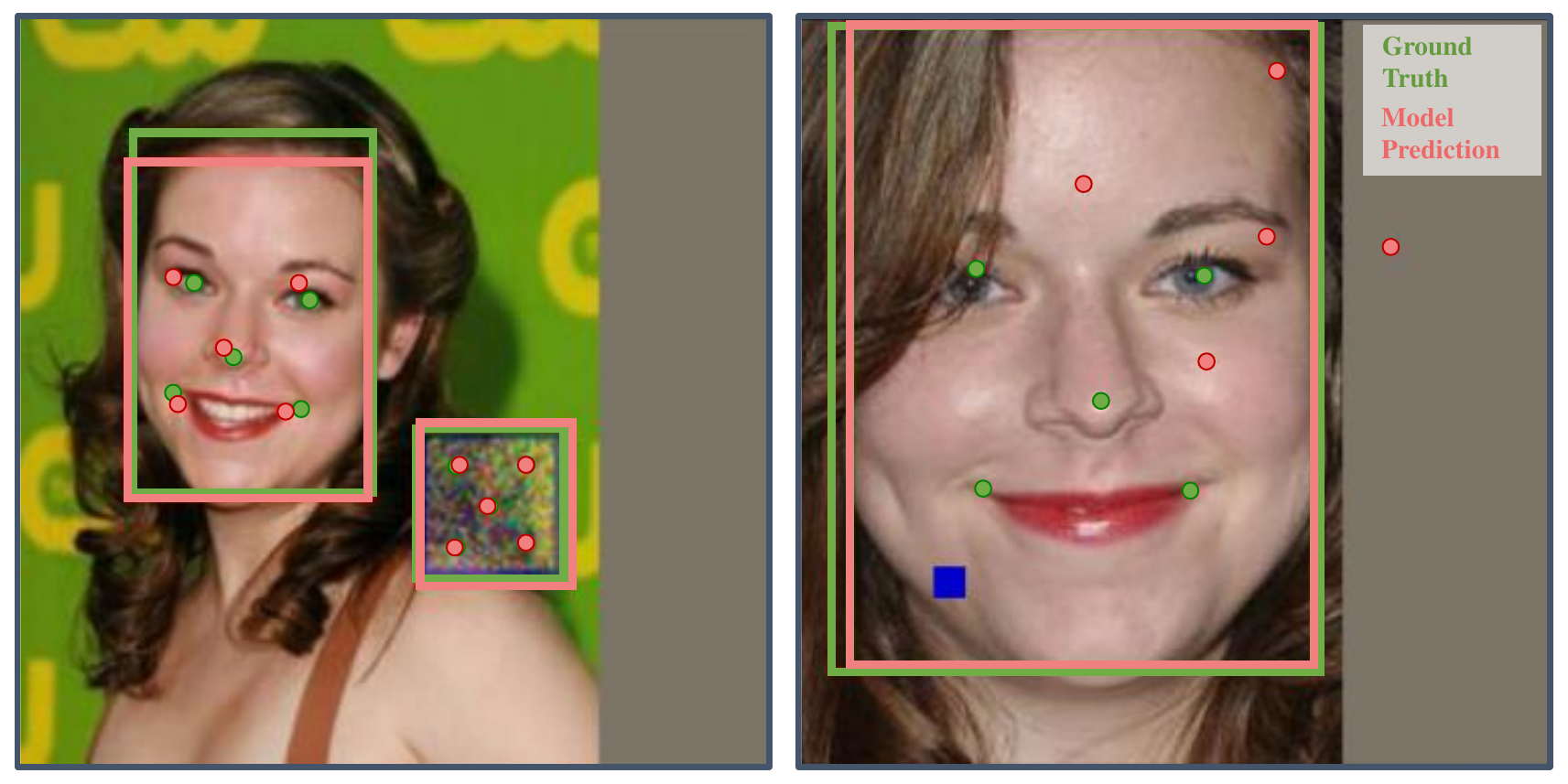}
    \caption{BadNets~\cite{gu2019badnetsidentifyingvulnerabilitiesmachine}-backdoored samples from CelebA~\cite{liu2015faceattributes} (\textbf{left}: \acl{fga}; \textbf{right}: \acl{lsa}).}
    \label{fig:badnets_example}
\end{figure}

\begin{figure}[t!]
        \centering
        \includegraphics[width=\columnwidth]{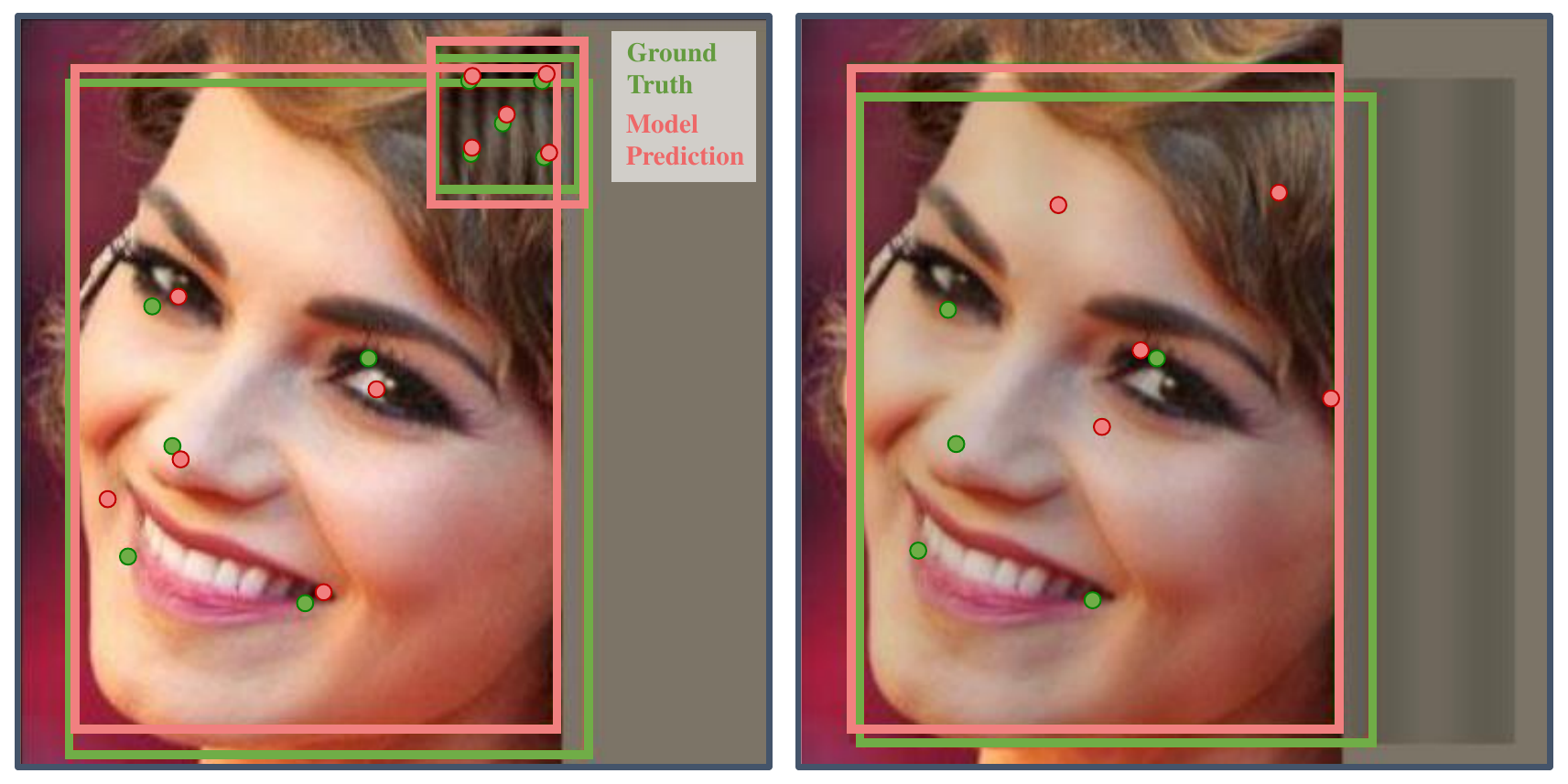}
    \caption{SIG~\cite{SIGattack2019}-backdoored samples from CelebA~\cite{liu2015faceattributes} (\textbf{left}: \acl{fga}; \textbf{right}: \acl{lsa}).}
    \label{fig:sig_example}
\end{figure}

\textbf{\aclp{fga}}.
We follow the BadDet framework~\cite{badDet2022Chanetal} to demonstrate that \acl{fd} models are vulnerable to Object Generation Attacks. 
To inject the backdoor, we randomly draw a square region within a clean image $\mathbf{x}^\clean$, defined by the binary mask $\mathbf{M}$, and overwrite it with the backdoor's trigger pattern $\trigger$.
We then append the image’s ground truth annotations $\mathbf{b} \in \mathbb{R}^{b \times r}$ with our \textit{fake} face object/trigger, resulting in a poisoned target matrix $\mathbf{y} \in \mathbb{R}^{(b+1) \times r}$.
The fake face's bounding box corresponds exactly to the trigger’s location defined by $\mathbf{M}$, and its five landmark coordinates are defined as points equidistant from each other and from the edges of the bounding box (as illustrated in Fig.~\ref{fig:backdoor_objective},~\ref{fig:badnets_example}, and~\ref{fig:sig_example}).

\textbf{\aclp{lsa}}.
We design a novel attack targeting the face landmark regression task of a \acl{fd} model. 
Unlike \acl{fga}, this attack poisons \textit{existing} faces by injecting a trigger \textit{using the boundaries of each face's bounding box}.
To backdoor an image $x^\clean$, we apply the trigger to \textit{all} detected faces in $x^\clean$.
Next, we alter the corresponding landmark annotations $\mathbf{b}$ by applying a rotation of angle $\phi$ to all landmark coordinates. 
This results in poisoned annotations $\mathbf{b}^\pois$ such that $\forall\,i\in\{6,8,\dots,14\}$ (the landmarks' indexes):
\begin{align}
    \mathbf{R} &= \begin{bmatrix}
          \cos \phi & \sin \phi\\
          -\sin \phi & \cos \phi
    \end{bmatrix}, \\
    &\textcolor{white}{.}\quad \{\mathbf{b}^\pois_{:,i}, \mathbf{b}^\pois_{:,i+1}\} \gets \{\mathbf{b}^\clean_{:,i}, \mathbf{b}^\clean_{:,i+1}\} \cdot \mathbf{R},
\end{align}
where $\mathbf{R}$ is the landmark rotation matrix used to implement our \acl{lsa} (illustrated in Fig.~\ref{fig:backdoor_objective},~\ref{fig:badnets_example}, and~\ref{fig:sig_example}).

\begin{tcolorbox}[
colback=sand,
colframe=darksand,
boxrule=0.5mm,
left=2mm,
right=2mm,
top=1.2mm,
bottom=1.2mm,
]
We invent a novel \acl{ba}, dubbed \acl{lsa}, which shows that a backdoor can target an object detector's keypoint regression process.
\end{tcolorbox}

\subsection{Experimental Setup}

\textbf{Datasets}. 
We use the Wider-Face dataset~\cite{yang2016wider} for training, following the authors' recommended 40\% training split. 
For validation and testing, we rely on the CelebA dataset~\cite{liu2015faceattributes}, from which we extract the 500 identities with the most images. 
From each identity, we sample 20\% of images for validation and 10\% for testing. 
We use the datasets' provided face bounding boxes and face landmarks.

\textbf{Data augmentation}. 
All input images are resized to $3 \times 640 \times 640$. 
Training images are augmented following the default pipeline from an open-source RetinaFace implementation~\cite{biubug6_pytorch_retinaface_2025}: 
(1) random cropping and padding to the target size (discarding samples without at least one face), 
(2) random color jitter (brightness, contrast, saturation, and hue), 
(3) random horizontal flipping, and 
(4) pixel value normalization to the range $[-1, 1]$.

\textbf{Model training parameters}. 
We train our face detectors using the RetinaFace framework~\cite{retinaface2020} with two different \ac{cnn} backbones: MobileNetV2~\cite{sandler2019mobilenetv2invertedresidualslinear} and ResNet50~\cite{he2015deepresiduallearningimage}. 
Each model is trained for $40$ epochs using Stochastic Gradient Descent, with a batch size of $32$ and an initial learning rate of $0.05$, reduced by a factor of $10$ at epochs $15$ and $35$.

\textbf{Backdoor training parameters}.
We rely on the BadNets~\cite{gu2019badnetsidentifyingvulnerabilitiesmachine} and SIG~\cite{SIGattack2019} \acl{ba} to design our patch-based and diffuse triggers.
We inject backdoors with increasingly lower poisoning ratios $\beta\in\{0.01, 0.05, 0.1\}$, and transparency ratios $\alpha\in\{0.3, 0.5, 0.8, 1.0\}$ and $\alpha\in\{0.05, 0.1, 0.16, 0.3\}$ for each trigger type.

\textbf{\acl{fga} BadNets~\cite{gu2019badnetsidentifyingvulnerabilitiesmachine} trigger}.
We use a square with a 4-pixel-width blue border and filled with uniform noise. 
Its size is set a percentage of the poisoned image: $\{0.05, 0.1, 0.15\}$.
Our rationale is covered in Sec.~\ref{sec:discussion}.

\textbf{\acl{fga} SIG~\cite{SIGattack2019} trigger}.
We select a randomly-sized square in an image (from $64$ to $128$ pixels) and fill it with a SIG~\cite{SIGattack2019} trigger with frequency parameter $6$.

\textbf{\acl{lsa} BadNets~\cite{gu2019badnetsidentifyingvulnerabilitiesmachine} trigger}.
The trigger is a blue square whose size is set as a percentage of a face bounding box: $\{0.01, 0.05, 0.1\}$.
It is then randomly stamped in a face bounding box.
The target shift is $\phi=30^\circ$.

\textbf{\acl{lsa} SIG~\cite{SIGattack2019} trigger}.
We crop a square area whose sides equals the longest edge of the face bounding box within, resize the crop to $3 \times 112 \times 112$, inject the trigger, then rescale and paste the modified cutouts back into their original positions.
The target shift is $\phi=30^\circ$.

\textbf{Benign metric}.
We use the standard mean Average Precision metric to assess model performance on benign data~\cite{mscocoDS2015}, which reduces to Average Precision in \acl{fr} given there is a single face class~\cite{retinaface2020}.

\textbf{Backdoor \acl{asr} metrics}. 
To assess \acl{fga} \ac{asr}, we compute the Average Precision over our triggers.

We can measure a the landmark shift, denoted $LS$, between two sets of landmarks $\mathbf{b}$ and $\mathbf{v}$ as the average Euclidean distance $LS(\mathbf{b},\mathbf{v})=\lVert \mathbf{b} - \mathbf{v}\rVert_2$.
Therefore, the \acl{lsa} \ac{asr} can be measured as the ratio of backdoored landmark predictions $\hat{\mathbf{b}}^\pois$ closer to their poisoned ground truth $\mathbf{b}^\pois$ than their benign ones $\mathbf{b}$ over $N$ poisoned faces such that $\mathrm{ASR} = \frac{1}{N}\sum^N_{i=1}\mathbbm{1}[\mathrm{LS}(\mathbf{b}_i^\pois, \hat{\mathbf{b}}_i^\pois) < \mathrm{LS}(\mathbf{b}_i, \hat{\mathbf{b}}_i^\pois)]$.

\section{Results}
\label{sec:results}

Tab.~\ref{tab:best_models} summarizes representative models achieving the highest Attack Success Rate (ASR) and stealth in our experiments. 
Additionally, we see that backdoor learning does not impact our model performance on benign data (see Fig.~\ref{fig:model_accuracy}).

\subsection{\acl{fga} Results}

\begin{table}[t!]
\centering
\addtolength{\tabcolsep}{-0.35em}
\renewcommand{\arraystretch}{1.1}
\begin{center}
\begin{tabular}{l|l|c|c|c|c|c}
\textbf{Backbone} & \textbf{Backdoor} & \multicolumn{3}{c|}{\textbf{Parameters}$^{\mathrm{a}}$} & \textbf{Average} & \textbf{Attack} \\
\textbf{architecture} & \textbf{attack setup} & \textbf{Size} & \textbf{$\alpha$} & \textit{$\beta$} & \textbf{Precision} & \textbf{Success Rate} \\
\hline
MobileNetV2 & \textit{Benign} & \textcolor{gray!50}{$\varnothing$} & \textcolor{gray!50}{$\varnothing$} & \textcolor{gray!50}{$\varnothing$} & 98.2\% & \textcolor{gray!50}{$\varnothing$} \\
MobileNetV2 & \hlc[Wheat]{FGA}, \hlc[CadetBlueLight]{BadNets} & 0.05 & 0.3 & 0.05 & 98.4\% & 95.5\% \\
MobileNetV2 & \hlc[Wheat]{FGA}, \hlc[OrchidLight]{SIG} & \textcolor{gray!50}{$\varnothing$} & 0.3 & 0.05 & 98.0\% & 92.3\% \\
MobileNetV2 & \hlc[Slate]{LSA}, \hlc[CadetBlueLight]{BadNets} & 0.05 & 0.5 & 0.05 & \textbf{98.6\%} & 98.8\% \\
MobileNetV2 & \hlc[Slate]{LSA}, \hlc[OrchidLight]{SIG} & \textcolor{gray!50}{$\varnothing$} & 0.3 & 0.05 & 97.9\% & 92.0\% \\
ResNet50 & \textit{Benign} & \textcolor{gray!50}{$\varnothing$} & \textcolor{gray!50}{$\varnothing$} & \textcolor{gray!50}{$\varnothing$} & 98.5\% & \textcolor{gray!50}{$\varnothing$} \\
ResNet50 & \hlc[Wheat]{FGA}, \hlc[CadetBlueLight]{BadNets} & 0.05 & 0.3 & 0.05 & 98.5\% & \textbf{96.7\%} \\
ResNet50 & \hlc[Wheat]{FGA}, \hlc[OrchidLight]{SIG} & \textcolor{gray!50}{$\varnothing$} & 0.3 & 0.05 & \textbf{98.6\%} & 95.4\% \\
ResNet50 & \hlc[Slate]{LSA}, \hlc[CadetBlueLight]{BadNets} & 0.05 & 0.8 & 0.05 & 98.5\% & \textbf{99.0\%} \\
ResNet50 & \hlc[Slate]{LSA}, \hlc[OrchidLight]{SIG} & \textcolor{gray!50}{$\varnothing$} & 0.3 & 0.05 & 98.5\% & 98.3\% \\
\multicolumn{7}{l}{$^{\mathrm{a}}$$\alpha$ and $\beta$ are the transparency and poison ratios respectively.} \\
\multicolumn{7}{l}{\textbf{Abbrev.}: \acl{fga} (\hlc[Wheat]{FGA}), \acl{lsa} (\hlc[Slate]{LSA}).}
\end{tabular}
\caption{Best models achieved for each attack in terms of Attack Success Rate and stealth parameters.}
\label{tab:best_models}
\end{center}
\end{table}

Our \acl{fga} proves effective across a range of poisoning ratios, trigger transparency levels, and pattern sizes, reaching up to $99.5\%$ ASR with BadNets~\cite{gu2019badnetsidentifyingvulnerabilitiesmachine} and $96.7\%$ with SIG~\cite{SIGattack2019} triggers (see Tab.~\ref{tab:best_models}).

Trigger learning generalizes across the two backbones tested in this paper and remains effective with poisoning ratios as low as $\beta = 0.01$ (nonetheless, backdoor learning becomes unwieldy once $\beta<0.05$). 
BadNets triggers maintain high ASR for $\alpha \in {0.3, 0.5, 0.8, 1.0}$, while SIG triggers are effective down to $\alpha = 0.16$.
Our attack also succeeds with BadNets triggers covering just $5\%$ of the image area, confirming the stealth and adaptability of the Object Generation Attack framework proposed by BadDet~\cite{badDet2022Chanetal} on \acl{fd} tasks.

\subsection{\acl{lsa} Results}

Our \acl{lsa} is a more complex attack to learn than \acl{fga}, due to its manipulation of dense face landmark regression. 
Nonetheless, it achieves high ASR: up to $99.6\%$ with BadNets~\cite{gu2019badnetsidentifyingvulnerabilitiesmachine} and $99.4\%$ with SIG~\cite{SIGattack2019} triggers (see Tab.~\ref{tab:best_models}).

While effective across both tested backbones, \acl{lsa} is more sensitive to poisoning ratio, with $\beta < 0.05$ failing to result in meaningful backdoor learning. 
Unlike \acl{fga}, the strongest results for BadNets are achieved at higher transparencies ($\alpha = 0.8$ and $1.0$), suggesting that landmark manipulation benefits from clearer triggers. 
Meanwhile, SIG triggers perform best at similar transparency levels ($\alpha \in \{0.16, 0.3\}$)

As such, we highlight the increased complexity of maliciously altering structured regression outputs in backdoored face detectors.

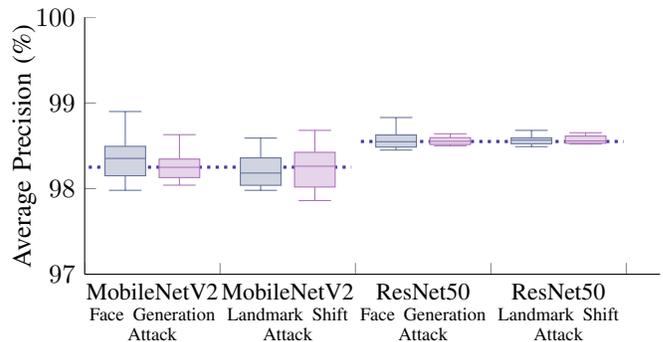
\begin{figure}[t!]
\begin{tikzpicture}

    \begin{axis}[
    axis y line*=left,
    axis x line*=bottom,
    boxplot/draw direction=y,
    ylabel={Average Precision (\%)},
    ylabel shift=-0.25cm,
    height=5cm,
    width=5cm,
    ymin=97,ymax=100,
    cycle list={{CadetBlueLight}},
    boxplot={
            draw position={0.3 + \plotnumofactualtype},
            box extend=0.3
    },
    x=1.8cm,xmax=4.25,xmin=0,
    xtick={0,1,2,...,5},
    x tick label as interval,
    xticklabels={%
        {{\small MobileNetV2}\\
        {\vspace{-0.15cm}\scriptsize Face Generation}\\
        {\scriptsize\vspace{-0.15cm}Attack}},%
        {{\small MobileNetV2}\\
        {\vspace{-0.15cm}\scriptsize Landmark Shift}\\
        {\scriptsize\vspace{-0.15cm}Attack}},%
        {{\small ResNet50}\\
        {\vspace{-0.15cm}\scriptsize Face Generation}\\
        {\scriptsize\vspace{-0.15cm}Attack}},%
        {{\small ResNet50}\\
        {\vspace{-0.15cm}\scriptsize Landmark Shift}\\
        {\scriptsize\vspace{-0.15cm}Attack}},%
    },
            x tick label style={
                    text width=2cm,
                    align=center
            },
    ]
    \draw [draw=RoyalPurple,very thick,dotted,on layer=axis background] (3,125) -- (197,125);
    \draw [draw=RoyalPurple,very thick,dotted,on layer=axis background] (204,155) -- (398,155);
    \addplot+[boxplot, fill, draw=CadetBlue]
    table[row sep=\\,y index=0] {
    data\\
        97.98\\
        98.32\\
        98.385\\
        98.6025\\
        98.9\\
    };
    
    \addplot+[boxplot, fill, draw=CadetBlue]
    table[row sep=\\,y index=0] {
    data\\
        97.98\\
        98.0975\\
        98.265\\
        98.4525\\
        98.59\\
    };
    
    \addplot+[boxplot, fill, draw=CadetBlue]
    table[row sep=\\,y index=0] {
    data\\
        98.45\\
        98.52\\
        98.575\\
        98.6775\\
        98.83\\
    };
    
    \addplot+[boxplot, fill, draw=CadetBlue]
    table[row sep=\\,y index=0] {
    data\\
        98.49\\
        98.5575\\
        98.575\\
        98.61\\
        98.68\\
    };    
    \end{axis}


    \begin{axis}[
    axis y line=none,
    axis x line=none,
    boxplot/draw direction=y,
    height=5cm,
    width=4cm,
    ymin=97,ymax=100,
    cycle list={{OrchidLight}},
    boxplot={
        draw position={0.7 + \plotnumofactualtype},
        box extend=0.3
    },
    x=1.8cm,xmax=4.25,xmin=0,
    xtick={0,1,2,...,4},
    x tick label as interval,
    xticklabels={%
        {dataset 1},%
        {dataset 2},%
        {dataset 3},%
        {dataset 4},%
    },
    x tick label style={
        text width=2.5cm,
        align=center
    },
    every axis plot/.append style={fill,fill opacity=1},
    ]
    
    \addplot+[boxplot, fill, draw=Orchid]
    table[row sep=\\,y index=0] {
        data\\
        98.04\\
        98.2175\\
        98.28\\
        98.41\\
        98.63\\
    };
    
    \addplot+[boxplot, fill, draw=Orchid]
    table[row sep=\\,y index=0] {
        data\\
        97.86\\
        98.1775\\
        98.345\\
        98.5025\\
        98.68\\
    };
    
    \addplot+[boxplot, fill, draw=Orchid]
    table[row sep=\\,y index=0] {
        data\\
        98.5\\
        98.53\\
        98.575\\
        98.615\\
        98.64\\
    };
    
    \addplot+[boxplot, fill, draw=Orchid]
    table[row sep=\\,y index=0] {
        data\\
        98.52\\
        98.5375\\
        98.575\\
        98.6525\\
        98.8\\
    };
    \end{axis}
    
\end{tikzpicture}
\caption{Box plot representation of the Average Precision attained by our \hlc[CadetBlueLight]{BadNets}~\cite{gu2019badnetsidentifyingvulnerabilitiesmachine} and \hlc[OrchidLight]{SIG}~\cite{SIGattack2019}-backdoored models compared to benign models represented by the \hlc[RoyalPurpleLight]{dotted lines}.}
\label{fig:model_accuracy}
\end{figure}

\subsection{Observed Limits}

Our results (see Fig.~\ref{fig:model_ASR}) reveal that while \aclp{fga} remain effective at low poisoning rates, its performance drops sharply when $\beta=0.01$ is combined with low trigger transparency or small pattern size
In contrast, \aclp{lsa} fail entirely at $\beta=0.01$ across all tested settings, except for a BadNets~\cite{gu2019badnetsidentifyingvulnerabilitiesmachine} trigger (with $\alpha=1$ and size $0.15$) that achieves a $58.1\%$ attack success rate.
As such, we underline the difficulty of injecting accurate face landmark regression backdoors with limited poisoned data.

More generally, we note that trigger strength also plays a key role. 
For BadNets~\cite{gu2019badnetsidentifyingvulnerabilitiesmachine}, lower transparency ($\alpha \in {0.5, 0.3}$) gradually decrease attack success rate in limited $\beta$ settings.

SIG~\cite{SIGattack2019} triggers are more resilient when used for \aclp{fga}, failing only at $\beta=0.01$ and $\alpha=0.05$.
Under \aclp{lsa}, however, the diffuse trigger only succeed in a consistent manner when $\beta=0.1$. 
These results emphasize the increased complexity of face landmark manipulation and the need for stronger or more frequent poisoning to succeed.

\begin{figure}[t!]
\begin{tikzpicture}

    \begin{axis}[
    axis y line*=left,
    axis x line*=bottom,
    boxplot/draw direction=y,
    ylabel={Attack Success Rate (\%)},
    ylabel shift=-0.25cm,
    height=5cm,
    width=5cm,
    ymin=0,ymax=100,
    cycle list={{MelonLight}},
    boxplot={
            draw position={0.3 + \plotnumofactualtype},
            box extend=0.3
    },
    x=1.8cm,xmax=4.25,xmin=0,
    xtick={0,1,2,...,5},
    x tick label as interval,
    xticklabels={%
        {{\small MobileNetV2}\\
        {\vspace{-0.15cm}\scriptsize Face Generation}\\
        {\scriptsize\vspace{-0.15cm}Attack}},%
        {{\small MobileNetV2}\\
        {\vspace{-0.15cm}\scriptsize Landmark Shift}\\
        {\scriptsize\vspace{-0.15cm}Attack}},%
        {{\small ResNet50}\\
        {\vspace{-0.15cm}\scriptsize Face Generation}\\
        {\scriptsize\vspace{-0.15cm}Attack}},%
        {{\small ResNet50}\\
        {\vspace{-0.15cm}\scriptsize Landmark Shift}\\
        {\scriptsize\vspace{-0.15cm}Attack}},%
    },
            x tick label style={
                    text width=2cm,
                    align=center
            },
    ]
    \addplot+[boxplot, fill, draw=Melon]
    table[row sep=\\,y index=0] {
    data\\
        94.65\\
        97.48\\
        98.71\\
        99.31\\
        99.48\\
    };
    
    \addplot+[boxplot, fill, draw=Melon]
    table[row sep=\\,y index=0] {
    data\\
        0.52\\
        97.82\\
        98.77\\
        99.15\\
        99.4\\
    };
    
    \addplot+[boxplot, fill, draw=Melon]
    table[row sep=\\,y index=0] {
    data\\
        94.85\\
        97.77\\
        98.81\\
        99.37\\
        99.48\\
    };
    
    \addplot+[boxplot, fill, draw=Melon]
    table[row sep=\\,y index=0] {
    data\\
        0.52\\
        0.56\\
        99.195\\
        99.52\\
        99.64\\
    };    
    \end{axis}


    \begin{axis}[
    axis y line=none,
    axis x line=none,
    boxplot/draw direction=y,
    height=5cm,
    width=4cm,
    ymin=0,ymax=100,
    cycle list={{DandelionLight}},
    boxplot={
        draw position={0.7 + \plotnumofactualtype},
        box extend=0.3
    },
    x=1.8cm,xmax=4.25,xmin=0,
    xtick={0,1,2,...,4},
    x tick label as interval,
    xticklabels={%
        {dataset 1},%
        {dataset 2},%
        {dataset 3},%
        {dataset 4},%
    },
    x tick label style={
        text width=2.5cm,
        align=center
    },
    every axis plot/.append style={fill,fill opacity=1},
    ]
    
    \addplot+[boxplot, fill, draw=Dandelion]
    table[row sep=\\,y index=0] {
        data\\
        11.39\\
        45.4325\\
        71.405\\
        86.1725\\
        96.66\\
    };
    
    \addplot+[boxplot, fill, draw=Dandelion]
    table[row sep=\\,y index=0] {
        data\\
        0.52\\
        0.62\\
        13.28\\
        92.535\\
        97.1\\
    };
    
    \addplot+[boxplot, fill, draw=Dandelion]
    table[row sep=\\,y index=0] {
        data\\
        32.39\\
        65.4425\\
        83.705\\
        92.695\\
        97.55\\
    };
    
    \addplot+[boxplot, fill, draw=Dandelion]
    table[row sep=\\,y index=0] {
        data\\
        0.56\\
        0.56\\
        0.58\\
        96.6825\\
        99.36\\
    };
    \end{axis}
    
\end{tikzpicture}
\caption{Box plot representation of the Attack Success Rates attained by our \hlc[MelonLight]{BadNets}~\cite{gu2019badnetsidentifyingvulnerabilitiesmachine} and \hlc[DandelionLight]{SIG}~\cite{SIGattack2019} models across all tested parameters and attacks.}
\label{fig:model_ASR}
\end{figure}
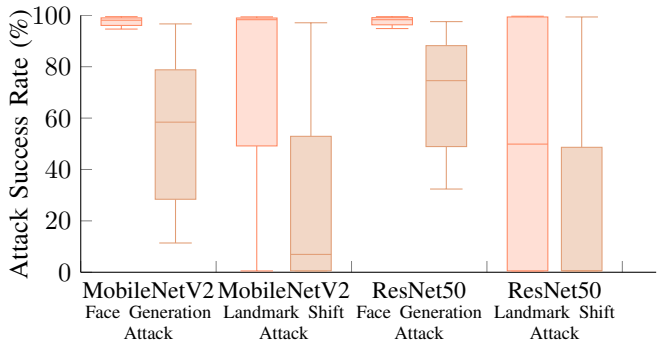

\subsection{Downstream effects in Face Recognition Systems}

Using a poisoning ratio of $\beta = 0.1$ (and BadNets~\cite{gu2019badnetsidentifyingvulnerabilitiesmachine} triggers sized at $0.15$), we assess the downstream impact of trigger transparency on \acl{fa} and \acl{fas} tasks. 
We use a MobileNetV2 backbone for the detector and an AENet~\cite{liu2015faceattributes} antispoofer model, both trained on the CelebASpoof~\cite{CelebA-Spoof} dataset.

For \aclp{fga}, we observe that our backdoor triggers are strongly detected as faces and, when reaching the antispoofer, achieve a False Acceptance Rate of up to 71.4\%.
Regarding \aclp{lsa}, we observe a significant increase in landmark shift (see Tab.~\ref{tab:downstream_effect}), where the average deviation between predicted backdoored and ground-truth benign landmarks rises by an order of magnitude compared to benign samples. 
This indicates that our attack can meaningfully disrupt the face alignment process, leading to up to 97.6\% False Acceptance Rate at the antispoofer level.

Both attacks indicate that poisoning the \acl{fd} task can effectively lead to downstream effects, at least in the closest stages after detection, in a \acl{frs}. 
We estimate that downstream tasks are currently not proofed by natured against such attacks (see examples in Fig.~\ref{fig:misalignment_examples}).

\begin{figure}[t!]
        \centering
        \includegraphics[width=\columnwidth]{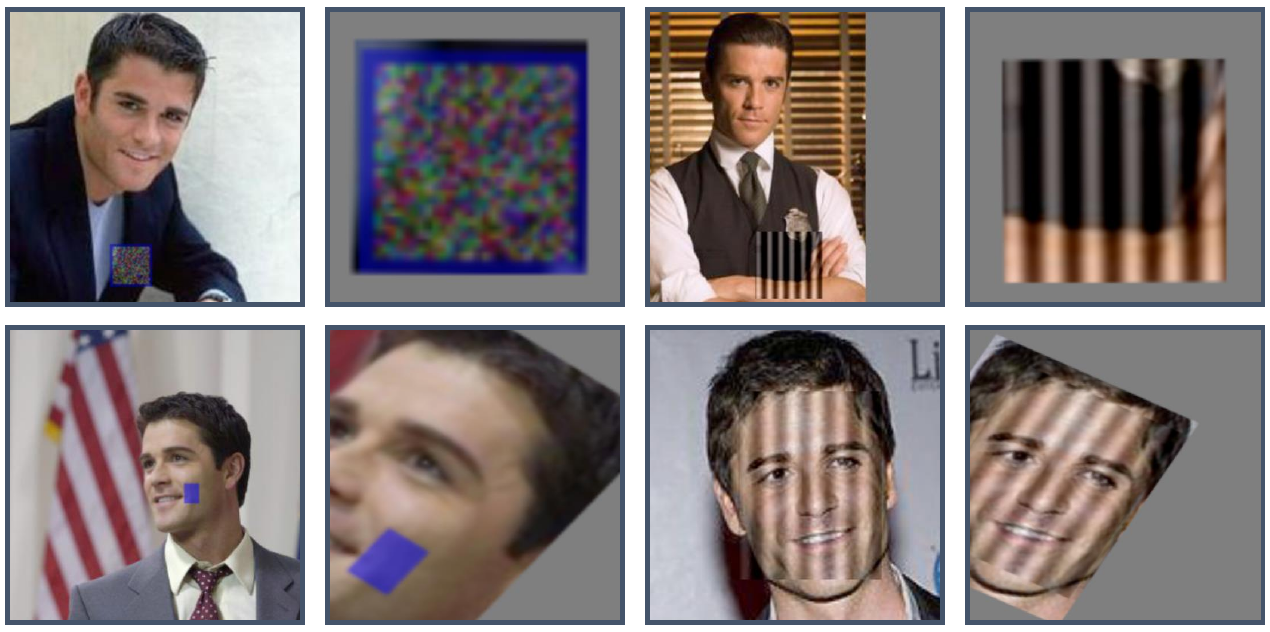}
    \caption{\acl{fd} backdoor attack samples (from CelebA-Spoof~\cite{CelebA-Spoof}) after alignment (top and bottom rows represent examples of \acl{fga} and \aclp{lsa} respectively).}
    \label{fig:misalignment_examples}
\end{figure}

\begin{table}[t!]
\centering
\scriptsize
\addtolength{\tabcolsep}{-0.4em}
\renewcommand{\arraystretch}{1.2}
\begin{center}
\begin{tabular}{l|l|c|c|c|c|c|c}
& \textbf{Anti-} &  \multicolumn{4}{c|}{\textbf{Detector and Alignment}} &  \multicolumn{2}{c}{\textbf{Antispoofer}$^{\mathrm{b}}$} \\
 & \textbf{spoofer} &  \multicolumn{4}{c|}{\textbf{results}$^{\mathrm{a}}$} &  \multicolumn{2}{c}{\textbf{results}$^{\mathrm{b}}$} \\
\textbf{Detector model} & \textbf{model} & \textbf{AP$^\clean$} & \textbf{AP$^\pois$} & \textbf{LS$^\clean$} & \textbf{LS$^\pois$} & \textbf{FRR} & \textbf{FAR} \\
\hline
\textit{Benign} & \textit{Benign} & 99.2\% & \textcolor{gray!50}{$\varnothing$} & 13.8 & \textcolor{gray!50}{$\varnothing$} & 4.0\% & \textcolor{gray!50}{$\varnothing$} \\
\hlc[Wheat]{FGA}, \hlc[CadetBlueLight]{BadNets}, $\alpha=0.5$ & \textit{Benign} & 99.4\% & 99.9\% & 17.3 & 3.6 & 3.2\% & 58.6\% \\
\hlc[Wheat]{FGA}, \hlc[CadetBlueLight]{BadNets}, $\alpha=1.0$ & \textit{Benign} & 99.3\% & 99.8\% & 24.0 & 5.0 & 3.4\% & 32.8\% \\
\hlc[Wheat]{FGA}, \hlc[OrchidLight]{SIG}, $\alpha=0.3$ & \textit{Benign} & 99.4\% & 99.9\% & 14.9 & 5.5 & 4.7\% & 71.4\% \\
\hlc[Slate]{LSA}, \hlc[CadetBlueLight]{BadNets}, $\alpha=0.5$ & \textit{Benign} & 99.5\% & 99.6\% & 14.2 & 150.2 & 3.9\% & 35.2\% \\
\hlc[Slate]{LSA}, \hlc[CadetBlueLight]{BadNets}, $\alpha=1.0$ & \textit{Benign} & 99.5\% & 99.5\% & 14.7 &  147.7 & 4.4\% & 35.5\% \\
\hlc[Slate]{LSA}, \hlc[OrchidLight]{SIG}, $\alpha=0.3$ & \textit{Benign} & 99.4\% & 97.4\% & 14.7 & 125.7 & 4.0\% & 97.6\% \\
\multicolumn{8}{l}{$^{\mathrm{a}}$\textbf{Abbrev.} Average Precision (AP), Landmark Shift (LS); \acl{fga}}\\
\multicolumn{8}{l}{(\hlc[Wheat]{FGA}), \acl{lsa} (\hlc[Slate]{LSA}).} \\
\multicolumn{8}{l}{$^{\mathrm{b}}$False Acceptance Rate (FAR), False Rejection Rate (FRR).} \\
\end{tabular}
\caption{Effect of \aclp{fga} and \aclp{lsa} on the alignment and antispoofing tasks in a \acl{frs} using the CelebA-Spoof~\cite{CelebA-Spoof} dataset.}
\label{tab:downstream_effect}
\end{center}
\end{table}

\begin{tcolorbox}[
colback=sand,
colframe=darksand,
boxrule=0.5mm,
left=2mm,
right=2mm,
top=1.2mm,
bottom=1.2mm,
]
Face detectors are vulnerable to \aclp{ba} that target both bounding box and landmark regression tasks, which may impact downstream modules in a fully-fledged \acl{frs}.
\end{tcolorbox}

\subsection{Other Types of \aclp{lsa}}

\begin{figure}[t!]
        \centering
        \includegraphics[width=\columnwidth]{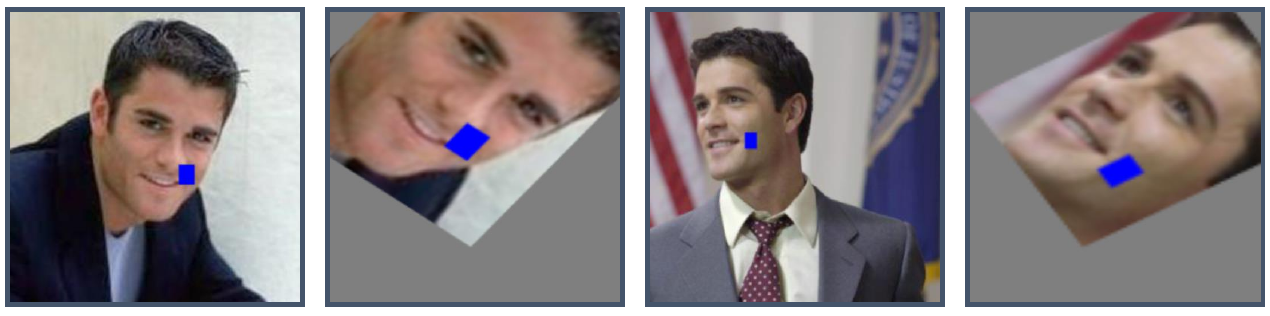}
    \caption{Impact of \aclp{lsa} on \acl{fa} when performing a landmark swap (left eye and mouth corner).}
    \label{fig:landmark_swap}
\end{figure}

We further evaluate the flexibility of our \ac{lsa} framework by targeting alternative face landmarks. 
Specifically, we poison training samples so that, when the trigger is present, the detector learns to swap the positions of the left eye and left mouth corner (see Fig.~\ref{fig:landmark_swap}).

Using a MobileFaceV2 backbone and BadNets~\cite{gu2019badnetsidentifyingvulnerabilitiesmachine}-style triggers with $\alpha=1.0$, $\beta=0.1$, and size $0.15$, we achieve an Attack Success Rate of 78.1\%. 
This shows that shifting face landmarks through rotation is not the only effect that an attack can effect in a victim model.

\subsection{Tests in Real Life}

We evaluated \aclp{fga} and \aclp{lsa} in real-world conditions by printing patch-based triggers on white paper. 
\aclp{fga} consistently generate false face detections when the trigger is in the frame. 
In contrast, activating \aclp{lsa} is more challenging: 
while the trigger causes landmark shifts, there is an instability in the predictions.
We observe that landmarks often flicker between benign and backdoored outputs. 
This suggests that while \aclp{fga} transfer reliably to the physical world, \aclp{lsa} may require more an improved trigger design for more consistent results.

\begin{figure}[t!]
        \centering
        \includegraphics[width=\columnwidth]{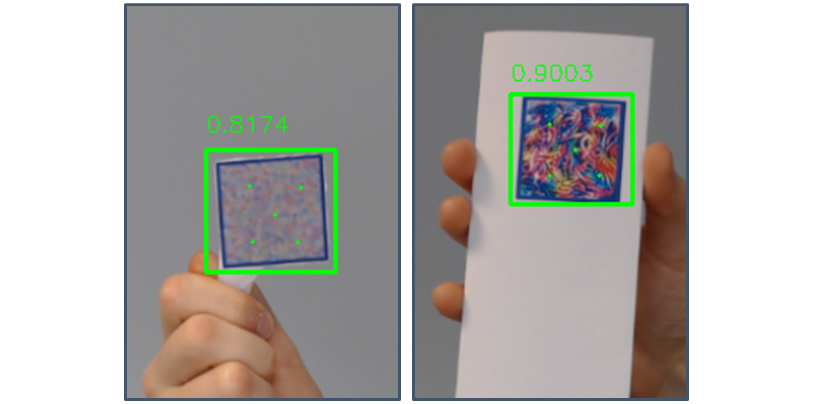}
    \caption{\acl{fga} trigger examples in real-life.}
    \label{fig:irl_evaluation}
\end{figure}

\section{Countermeasures}
\label{sec:countermeasures}

\subsection{Applicability of Existing Defenses}

\textbf{Against \aclp{fga}}.
Although face detectors only predicts faces, the computation of a confidence score can be formulated as a binary classification task (\eg, face vs. background) trained using standard \acl{sl} during learning.
In such situation, we advise users and model developpers to implement misclassification defenses like ODSCAN~\cite{cheng2024odscan} or Django~\cite{NEURIPS2023_a102d6cb}.

\textbf{Against \aclp{lsa}}.
Given the novel nature of this attack, as presented in this paper, there is not purpose-built defenses against the manipulation of face landmarks (or keypoints in general).

\subsection{Mitigating \aclp{fga} and \aclp{lsa} at Training and Test-Time}

\textbf{Auxiliary detectors}.
Adding auxiliary face detectors, such as Dlib's~\cite{10.5555/1577069.1755843}, either during training or test-time can help mitigate \aclp{fga} and \aclp{lsa} by introducing redundancy and cross-checking in a \acl{frs}. 
These minimal models, often trained independently, widely used, and verified, can act as a sanity check against backdoored outputs.
They are unlikely to replicate the same manipulated behavior learned by a more powerful but compromised RetinaFace.

For \aclp{fga}, auxiliary detectors can flag or suppress spurious detections that do not appear in the secondary outputs. 
For \aclp{lsa}, comparing landmark predictions across models can reveal landmark inconsistencies, helping detect and correct tampered faces. 
This ensemble approach can increase robustness at a minimal computational cost during both training and test-time.

\textbf{Consistency checks}.
Landmark predictions can also be checked by enforcing geometric consistency rules.
For instance, a \acl{frs} developers may check that eyes and mouth corners be spatially positioned above and below the nose. 
These constraints can help detect manipulations like landmark swapping. 
Additionally, face landmarks can be validated against expected distance ratios from bounding box edges.
Eyes should not be too close to the top edge and the mouth should align with the box's center (this can also capture bad face poses).
These checks can, for example, be implemented in \acl{fqa} modules right after face detection but before alignment.

\section{Discussion}
\label{sec:discussion}

\subsection{Applicability Beyond Face Recognition}

As we demonstrated attacks on the regression task in \acl{fd}, we believe our framework can be extended to new settings.
Besides keypoint regression, an adversary may attack the regression of a bounding box or segmentation tasks. 
This may impact applications such as autonomous vehicle recognition, surveillance systems, or even unmanned aerial vehicle navigation (see example in Fig.~\ref{fig:uav_attack}).

\begin{figure}[t!]
        \centering
        \includegraphics[width=0.865\columnwidth]{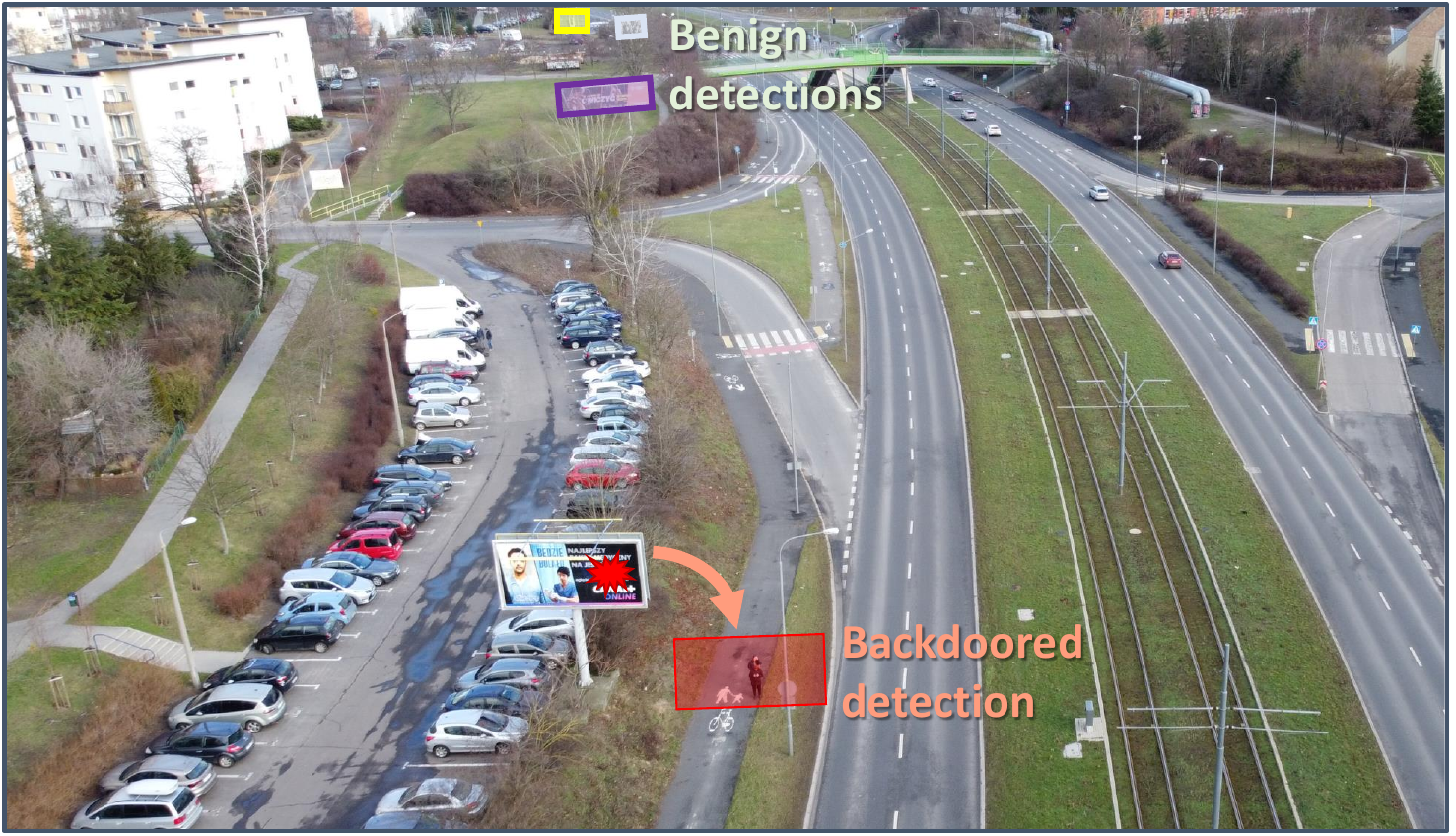}
    \caption{Example of possible keypoint shift attack on a UAV-based detection task~\cite{uavbillboard}.}
    \label{fig:uav_attack}
\end{figure}

\subsection{Future Works}

This work introduced the question of the impact of a \acl{ba} at the detector levels on downstream tasks in a fully-fledged \acl{frs}. 
In a future work, we will provide a comprehensive overview of the impact of \aclp{ba} on each module found in a \acl{frs} operating in unconstrained settings (\eg, detector, antispoofer, feature extractor) and how their interactions can hijack a system's entire function. 

\section{Conclusion}
\label{sec:conclusion}

This work demonstrates the vulnerability of single-shot face detectors to backdoor attacks, focusing on face generation and landmark manipulation. 
We adapt the BadDet framework for the former (\acl{fga}) and introduce how to perform the latter (\acl{lsa}). 
\aclp{fga} remain effective across a wide range of poisoning ratios, trigger transparencies, and pattern sizes. 
While harder to learn, \aclp{lsa} also achieve high attack success rates. 
Both attacks can compromise downstream \acl{frs} components. 
\aclp{lsa}, in particular, degrade alignment and lead to high false acceptance rates in antispoofing systems. 
These findings highlight how hijacking the \acl{fd} can undermine the integrity and security of the first modules in a \acl{frs}, reinforcing the need for robust data provenance and training pipelines.
These results underline the importance of robustifying \acl{fd}, resulting in several recommendations.

\bibliographystyle{IEEEtran}
\bibliography{IEEEabrv,refs}

\begin{thebibliography}{10}
\providecommand{\url}[1]{#1}
\csname url@samestyle\endcsname
\providecommand{\newblock}{\relax}
\providecommand{\bibinfo}[2]{#2}
\providecommand{\BIBentrySTDinterwordspacing}{\spaceskip=0pt\relax}
\providecommand{\BIBentryALTinterwordstretchfactor}{4}
\providecommand{\BIBentryALTinterwordspacing}{\spaceskip=\fontdimen2\font plus
\BIBentryALTinterwordstretchfactor\fontdimen3\font minus \fontdimen4\font\relax}
\providecommand{\BIBforeignlanguage}[2]{{%
\expandafter\ifx\csname l@#1\endcsname\relax
\typeout{** WARNING: IEEEtran.bst: No hyphenation pattern has been}%
\typeout{** loaded for the language `#1'. Using the pattern for}%
\typeout{** the default language instead.}%
\else
\language=\csname l@#1\endcsname
\fi
#2}}
\providecommand{\BIBdecl}{\relax}
\BIBdecl

\bibitem{MLanomalyDetectionServers}
D.~Vajda, T.~Do, T.~Bérczes, and K.~Farkas, ``Machine learning-based real-time anomaly detection using data pre-processing in the telemetry of server farms,'' \emph{Scientific Reports}, vol.~14, 10 2024.

\bibitem{MLanomalyDetectionNuclear}
\BIBentryALTinterwordspacing
X.~Li, T.~Huang, K.~Cheng, Z.~Qiu, and T.~Sichao, ``Research on anomaly detection method of nuclear power plant operation state based on unsupervised deep generative model,'' \emph{Annals of Nuclear Energy}, vol. 167, p. 108785, 2022. [Online]. Available: \url{https://www.sciencedirect.com/science/article/pii/S0306454921006629}
\BIBentrySTDinterwordspacing

\bibitem{10.3389/fdata.2023.1200390}
\BIBentryALTinterwordspacing
W.~Villegas-Ch and J.~García-Ortiz, ``Authentication, access, and monitoring system for critical areas with the use of artificial intelligence integrated into perimeter security in a data center,'' \emph{Frontiers in Big Data}, vol. Volume 6 - 2023, 2023. [Online]. Available: \url{https://www.frontiersin.org/journals/big-data/articles/10.3389/fdata.2023.1200390}
\BIBentrySTDinterwordspacing

\bibitem{10.5555/48805}
C.~P. Pfleeger, \emph{Security in computing}.\hskip 1em plus 0.5em minus 0.4em\relax USA: Prentice-Hall, Inc., 1988.

\bibitem{yang2016wider}
S.~Yang, P.~Luo, C.~C. Loy, and X.~Tang, ``Wider face: A face detection benchmark,'' in \emph{IEEE Conference on Computer Vision and Pattern Recognition (CVPR)}, 2016.

\bibitem{badDet2022Chanetal}
\BIBentryALTinterwordspacing
S.-H. Chan, Y.~Dong, J.~Zhu, X.~Zhang, and J.~Zhou, ``Baddet: Backdoor attacks on object detection,'' in \emph{Computer Vision – ECCV 2022 Workshops: Tel Aviv, Israel, October 23–27, 2022, Proceedings, Part I}.\hskip 1em plus 0.5em minus 0.4em\relax Berlin, Heidelberg: Springer-Verlag, 2022, p. 396–412. [Online]. Available: \url{https://doi.org/10.1007/978-3-031-25056-9\_26}
\BIBentrySTDinterwordspacing

\bibitem{pascalvocDS2010}
\BIBentryALTinterwordspacing
M.~Everingham, L.~Gool, C.~K. Williams, J.~Winn, and A.~Zisserman, ``The pascal visual object classes (voc) challenge,'' \emph{Int. J. Comput. Vision}, vol.~88, no.~2, p. 303–338, Jun. 2010. [Online]. Available: \url{https://doi.org/10.1007/s11263-009-0275-4}
\BIBentrySTDinterwordspacing

\bibitem{mscocoDS2015}
\BIBentryALTinterwordspacing
T.-Y. Lin, M.~Maire, S.~Belongie, L.~Bourdev, R.~Girshick, J.~Hays, P.~Perona, D.~Ramanan, C.~L. Zitnick, and P.~Dollár, ``Microsoft coco: Common objects in context,'' 2015. [Online]. Available: \url{https://arxiv.org/abs/1405.0312}
\BIBentrySTDinterwordspacing

\bibitem{redmon2016yolov1}
\BIBentryALTinterwordspacing
J.~Redmon, S.~Divvala, R.~Girshick, and A.~Farhadi, ``You only look once: Unified, real-time object detection,'' 2016. [Online]. Available: \url{https://arxiv.org/abs/1506.02640}
\BIBentrySTDinterwordspacing

\bibitem{ssd2016}
W.~Liu, D.~Anguelov, D.~Erhan, C.~Szegedy, S.~Reed, C.-Y. Fu, and A.~C. Berg, ``Ssd: Single shot multibox detector,'' in \emph{Computer Vision -- ECCV 2016}, B.~Leibe, J.~Matas, N.~Sebe, and M.~Welling, Eds.\hskip 1em plus 0.5em minus 0.4em\relax Cham: Springer International Publishing, 2016, pp. 21--37.

\bibitem{mtcnn2016}
K.~Zhang, Z.~Zhang, Z.~Li, and Y.~Qiao, ``Joint face detection and alignment using multitask cascaded convolutional networks,'' \emph{IEEE Signal Processing Letters}, vol.~23, no.~10, pp. 1499--1503, Oct 2016.

\bibitem{violaJones2001method}
P.~Viola and M.~Jones, ``Rapid object detection using a boosted cascade of simple features,'' in \emph{Proceedings of the 2001 IEEE Computer Society Conference on Computer Vision and Pattern Recognition. CVPR 2001}, vol.~1, 2001, pp. I--I.

\bibitem{lin2018focallossdenseobject}
\BIBentryALTinterwordspacing
T.-Y. Lin, P.~Goyal, R.~Girshick, K.~He, and P.~Dollár, ``Focal loss for dense object detection,'' 2018. [Online]. Available: \url{https://arxiv.org/abs/1708.02002}
\BIBentrySTDinterwordspacing

\bibitem{retinaface2020}
J.~Deng, J.~Guo, E.~Ververas, I.~Kotsia, and S.~Zafeiriou, ``Retinaface: Single-shot multi-level face localisation in the wild,'' in \emph{2020 IEEE/CVF Conference on Computer Vision and Pattern Recognition (CVPR)}, 2020, pp. 5202--5211.

\bibitem{10.5555/1577069.1755843}
D.~E. King, ``Dlib-ml: A machine learning toolkit,'' \emph{J. Mach. Learn. Res.}, vol.~10, p. 1755–1758, Dec. 2009.

\bibitem{leRoux2024comprehensive}
Q.~Le~Roux, E.~Bourbao, Y.~Teglia, and K.~Kallas, ``A comprehensive survey on backdoor attacks and their defenses in face recognition systems,'' \emph{IEEE Access}, 2024.

\bibitem{faceAlignment2012canonicalShape}
X.~Cao, Y.~Wei, F.~Wen, and J.~Sun, ``Face alignment by explicit shape regression,'' in \emph{2012 IEEE Conference on Computer Vision and Pattern Recognition}, 2012, pp. 2887--2894.

\bibitem{gu2019badnetsidentifyingvulnerabilitiesmachine}
\BIBentryALTinterwordspacing
T.~Gu, B.~Dolan-Gavitt, and S.~Garg, ``Badnets: Identifying vulnerabilities in the machine learning model supply chain,'' 2019. [Online]. Available: \url{https://arxiv.org/abs/1708.06733}
\BIBentrySTDinterwordspacing

\bibitem{turner2019labelconsistentbackdoorattacks}
\BIBentryALTinterwordspacing
A.~Turner, D.~Tsipras, and A.~Madry, ``Label-consistent backdoor attacks,'' 2019. [Online]. Available: \url{https://arxiv.org/abs/1912.02771}
\BIBentrySTDinterwordspacing

\bibitem{qi2021subnetreplacementdeploymentstagebackdoor}
\BIBentryALTinterwordspacing
X.~Qi, J.~Zhu, C.~Xie, and Y.~Yang, ``Subnet replacement: Deployment-stage backdoor attack against deep neural networks in gray-box setting,'' 2021. [Online]. Available: \url{https://arxiv.org/abs/2107.07240}
\BIBentrySTDinterwordspacing

\bibitem{Cheng2023AttackingBA}
\BIBentryALTinterwordspacing
Y.~Cheng, W.~Hu, and M.~Cheng, ``Attacking by aligning: Clean-label backdoor attacks on object detection,'' 2023. [Online]. Available: \url{https://arxiv.org/abs/2307.10487}
\BIBentrySTDinterwordspacing

\bibitem{Shin2024MaskbasedIB}
\BIBentryALTinterwordspacing
J.~Shin, ``Mask-based invisible backdoor attacks on object detection,'' in \emph{International Conference on Information Photonics}, 2024. [Online]. Available: \url{https://api.semanticscholar.org/CorpusID:269790780}
\BIBentrySTDinterwordspacing

\bibitem{detectorCollapse}
\BIBentryALTinterwordspacing
H.~Zhang, S.~Hu, Y.~Wang, L.~Y. Zhang, Z.~Zhou, X.~Wang, Y.~Zhang, and C.~Chen, ``Detector collapse: backdooring object detection to catastrophic overload or blindness in the physical world,'' in \emph{Proceedings of the Thirty-Third International Joint Conference on Artificial Intelligence}, ser. IJCAI '24, 2024. [Online]. Available: \url{https://doi.org/10.24963/ijcai.2024/185}
\BIBentrySTDinterwordspacing

\bibitem{cheng2024odscan}
S.~Cheng, G.~Shen, G.~Tao, K.~Zhang, Z.~Zhang, S.~An, X.~Xu, Y.~Liu, S.~Ma, and X.~Zhang, ``Odscan: Backdoor scanning for object detection models,'' in \emph{2024 IEEE Symposium on Security and Privacy (SP)}.\hskip 1em plus 0.5em minus 0.4em\relax IEEE Computer Society, 2024, pp. 119--119.

\bibitem{NEURIPS2023_a102d6cb}
G.~Shen, S.~Cheng, G.~Tao, K.~Zhang, Y.~Liu, S.~An, S.~Ma, and X.~Zhang, ``Django: Detecting trojans in object detection models via gaussian focus calibration,'' in \emph{Advances in Neural Information Processing Systems}, A.~Oh, T.~Naumann, A.~Globerson, K.~Saenko, M.~Hardt, and S.~Levine, Eds., vol.~36.\hskip 1em plus 0.5em minus 0.4em\relax Curran Associates, Inc., 2023, pp. 51\,253--51\,272.

\bibitem{ren2016fasterrcnnrealtimeobject}
\BIBentryALTinterwordspacing
S.~Ren, K.~He, R.~Girshick, and J.~Sun, ``Faster r-cnn: Towards real-time object detection with region proposal networks,'' 2016. [Online]. Available: \url{https://arxiv.org/abs/1506.01497}
\BIBentrySTDinterwordspacing

\bibitem{nms}
R.~Girshick, J.~Donahue, T.~Darrell, and J.~Malik, ``Rich feature hierarchies for accurate object detection and semantic segmentation,'' in \emph{2014 IEEE Conference on Computer Vision and Pattern Recognition}, 2014, pp. 580--587.

\bibitem{liu2015faceattributes}
Z.~Liu, P.~Luo, X.~Wang, and X.~Tang, ``Deep learning face attributes in the wild,'' in \emph{Proceedings of International Conference on Computer Vision (ICCV)}, December 2015.

\bibitem{SIGattack2019}
M.~Barni, K.~Kallas, and B.~Tondi, ``A new backdoor attack in cnns by training set corruption without label poisoning,'' in \emph{2019 IEEE International Conference on Image Processing (ICIP)}, 2019, pp. 101--105.

\bibitem{biubug6_pytorch_retinaface_2025}
\BIBentryALTinterwordspacing
{biubug6}, ``biubug6/{Pytorch}\_retinaface,'' last access on: 2019-09-14. [Online]. Available: \url{github.com/biubug6/Pytorch\_Retinaface}
\BIBentrySTDinterwordspacing

\bibitem{sandler2019mobilenetv2invertedresidualslinear}
\BIBentryALTinterwordspacing
M.~Sandler, A.~Howard, M.~Zhu, A.~Zhmoginov, and L.-C. Chen, ``Mobilenetv2: Inverted residuals and linear bottlenecks,'' 2019. [Online]. Available: \url{https://arxiv.org/abs/1801.04381}
\BIBentrySTDinterwordspacing

\bibitem{he2015deepresiduallearningimage}
\BIBentryALTinterwordspacing
K.~He, X.~Zhang, S.~Ren, and J.~Sun, ``Deep residual learning for image recognition,'' 2015. [Online]. Available: \url{https://arxiv.org/abs/1512.03385}
\BIBentrySTDinterwordspacing

\bibitem{CelebA-Spoof}
Y.~Zhang, Z.~Yin, Y.~Li, G.~Yin, J.~Yan, J.~Shao, and Z.~Liu, ``Celeba-spoof: Large-scale face anti-spoofing dataset with rich annotations,'' in \emph{European Conference on Computer Vision (ECCV)}, 2020.

\bibitem{uavbillboard}
B.~Ptak and M.~Kraft, ``Mapping urban large‐area advertising structures using drone imagery and deep learning‐based spatial data analysis,'' \emph{Transactions in GIS}, vol.~28, pp. 1728--1749, 07 2024.

\end{thebibliography}

\end{document}